\documentclass[10pt,twocolumn,letterpaper]{article}

\usepackage[]{cvpr}
\usepackage{booktabs}
\usepackage{tabularx}
\usepackage{enumitem}
\usepackage{algorithm}
\usepackage{algpseudocode}
\usepackage[]{todonotes}
\usepackage[accsupp]{axessibility}

\usepackage{graphicx}
\usepackage{tikz}
\usetikzlibrary{shapes.geometric, arrows, arrows.meta, positioning}
\usepackage{caption}

\usepackage[normalem]{ulem}  
\usepackage{xcolor}

\definecolor{cvprblue}{rgb}{0.21,0.49,0.74}
\usepackage[pagebackref,breaklinks,colorlinks,allcolors=cvprblue]{hyperref}
\usepackage{orcidlink} 

\newcommand{\orcid}[1]{\orcidlink{#1}}

\usepackage{listings}

\definecolor{codegreen}{rgb}{0,0.6,0}
\definecolor{codegray}{rgb}{0.5,0.5,0.5}
\definecolor{codepurple}{rgb}{0.58,0,0.82}
\definecolor{backcolour}{rgb}{0.98,0.98,0.98}

\lstdefinestyle{mystyle}{
    backgroundcolor=\color{backcolour},   
    commentstyle=\color{codegreen},
    keywordstyle=\color{magenta},
    numberstyle=\tiny\color{codegray},
    stringstyle=\color{codepurple},
    basicstyle=\ttfamily\small,
    breakatwhitespace=false,         
    breaklines=true,                 
    captionpos=b,                    
    keepspaces=true,                 
    numbers=none,                    
    numbersep=10pt,                  
    showspaces=false,                
    showstringspaces=false,
    showtabs=false,                  
    tabsize=2,
    frame=tb,
    title=\lstname
}

\renewcommand{\thesection}{\Alph{section}}

\usepackage{microtype}

\definecolor{cvprblue}{rgb}{0.21,0.49,0.74}
\usepackage[pagebackref,breaklinks,colorlinks,allcolors=cvprblue]{hyperref}

\title{Towards Context-Aware Image Anonymization with Multi-Agent Reasoning}

\author{
Robert Aufschl\"{a}ger \orcid{0009-0004-0986-3504} \quad
Jakob Folz \orcid{0000-0002-8428-1916} \quad
Gautam Savaliya \orcid{0009-0008-8400-6368} \\
Manjitha D Vidanalage \orcid{0009-0006-0395-651X} \quad
Michael Heigl \orcid{0000-0001-7303-113X} \quad
Martin Schramm \orcid{0000-0001-6206-2969} \\[2mm]
Deggendorf Institute of Technology, Deggendorf, Germany \\[1mm]
{\tt\small robert.aufschlaeger@th-deg.de}
}

\begin{document}

\maketitle

\begin{abstract}
Street-level imagery contains personally identifiable information (PII), some of which is context-dependent. Existing anonymization methods either over-process images or miss subtle identifiers, while API-based solutions compromise data sovereignty.
We present an agentic framework CAIAMAR (\underline{C}ontext-\underline{A}ware \underline{I}mage \underline{A}nonymization with \underline{M}ulti-\underline{A}gent \underline{R}easoning) for context-aware PII segmentation with diffusion-based anonymization, combining pre-defined processing for high-confidence cases with multi-agent reasoning for indirect identifiers. Three specialized agents coordinate via round-robin speaker selection in a Plan-Do-Check-Act (PDCA) cycle, enabling large vision-language models to classify PII based on spatial context (private vs. public property) rather than rigid category rules.
The agents implement spatially-filtered coarse-to-fine detection where a scout-and-zoom strategy identifies candidates, open-vocabulary segmentation processes localized crops, and $IoU$-based deduplication ($30\%$ threshold) prevents redundant processing. Modal-specific diffusion guidance with appearance decorrelation substantially reduces re-identification (Re-ID) risks.
On CUHK03-NP, our method reduces person Re-ID risk by $73\%$ ($R1$: $16.9\%$ vs. $62.4\%$ baseline). For image quality preservation on CityScapes, we achieve KID: $0.001$, and FID: $9.1$, significantly outperforming existing anonymization. The agentic workflow detects non-direct PII instances across object categories, and downstream semantic segmentation is preserved.
Operating entirely on-premise with open-source models, the framework generates human-interpretable audit trails supporting EU's GDPR transparency requirements while flagging failed cases for human review.

\end{abstract}
\begin{figure}[t!]
    \centering
    \includegraphics[width=\columnwidth]{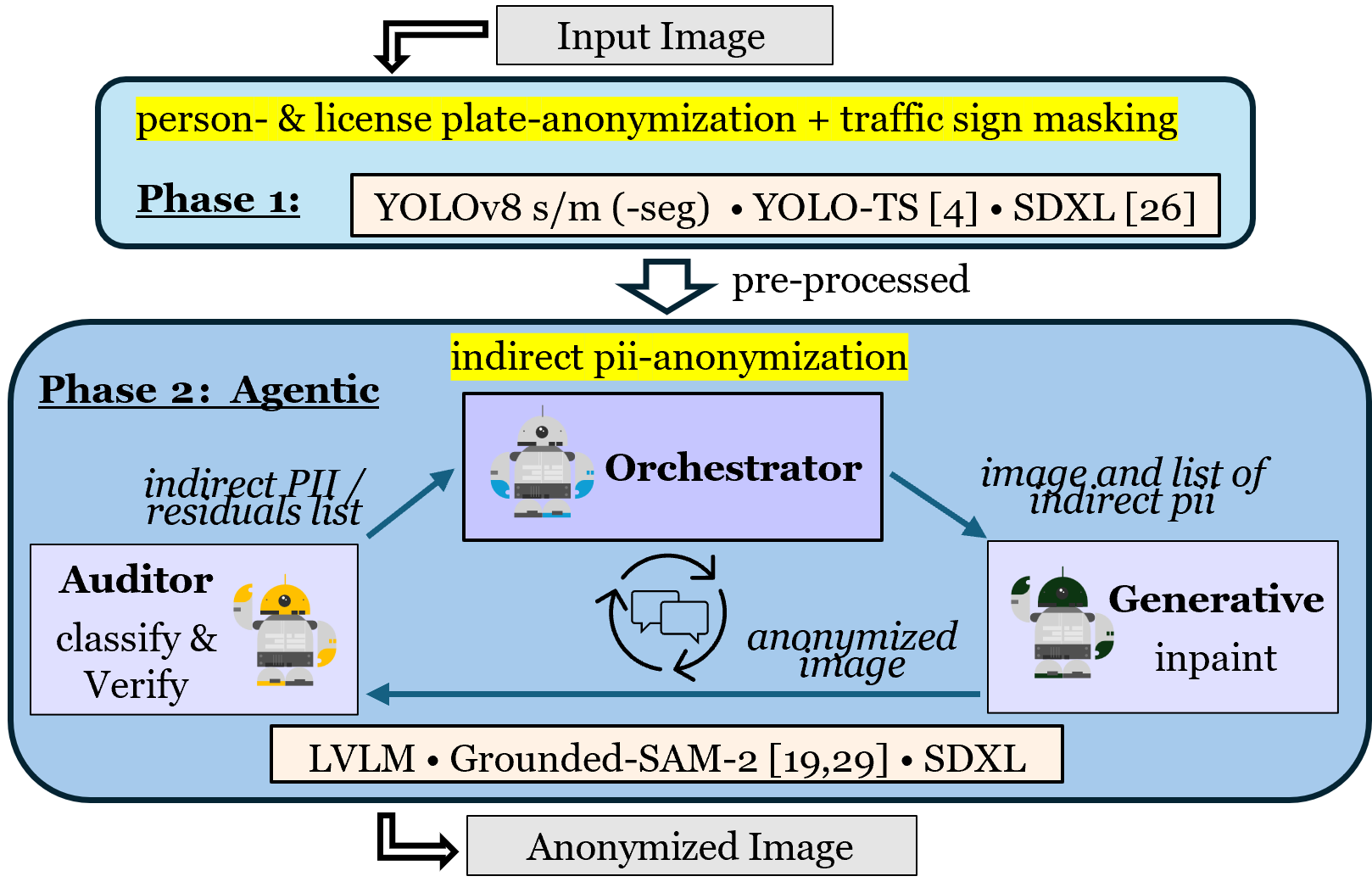}
    \caption{\textbf{Two-phase agentic anonymization architecture.} Phase 1 employs specialized models for direct PII (full body, license plates). Phase 2 implements multi-agent orchestration with \emph{round-robin} coordination, where specialized agents handle classification (Auditor), synthesis (Generative), and workflow management (Orchestrator), implementing PDCA cycles.}
    \label{fig:architecture}
\end{figure}

\section{Introduction}

Street-level imagery has become ubiquitous through mapping services, autonomous vehicle datasets, and urban monitoring systems. While enabling applications in navigation, urban planning, and computer vision research, these images inherently capture personally identifiable information (PII) at scale. Beyond obvious identifiers like pedestrian faces and license plates, recent work demonstrates that indirect identifiers, including clothing, accessories, and contextual objects, enable Re-ID and group membership inference even when direct PII is obscured~\cite{franchi2025, tomekcce2024private}.

The challenge is compounded by advances in large vision-language models (LVLMs). State-of-the-art models can infer private attributes from images without direct human depictions with up to $76.4\%$ accuracy~\cite{tomekcce2024private}, while cross-modal person re-identification (Re-ID) systems achieve robust performance across challenging cross-dataset and zero-shot scenarios~\cite{li2024all, aufschlaeger2025following}. Recent foundation model-based multimodal Re-ID frameworks can handle uncertain input modalities
achieving competitive zero-shot performance without fine-tuning~\cite{li2024all}. Existing solutions fail to address the full scope of this challenge. 
Monolithic detection pipelines apply uniform processing regardless of context, missing the distinction between public signage versus private property; black-box approaches provide no audit trail for regulatory compliance;
and most critically, current methods lack adaptive strategies for the semantic diversity of indirect identifiers. We propose that effective anonymization requires not just better models but a fundamentally different architecture: one that reasons about context, adapts to diverse PII types, and provides transparent decision-making. 
To this end, we introduce a multi-agent framework that orchestrates specialized components through structured communication and iterative refinement. Our framework addresses the \textbf{research question}: \textit{``Can orchestrated agents achieve context-aware image anonymization while preserving data utility and enabling accountability?''} 
The system (Figure~\ref{fig:architecture}) combines pre-defined preprocessing for high-confidence direct identifiers with agentic reasoning for context-dependent cases, implementing bounded iterative refinement through \emph{Plan–Do–Check–Act (PDCA)} cycles (Shewhart / Deming cycles) for multi-agent coordination. 

Our key contributions are:
\begin{enumerate}[leftmargin=*, itemsep=2pt, parsep=0pt]
    \item \textbf{Context-aware agentic anonymization with PDCA refinement.} 
    We present an image anonymization system combining pre-defined preprocessing for direct PII (full bodies, license plates) with multi-agent orchestration for context-dependent identifiers requiring semantic reasoning. The framework implements spatially-filtered coarse-to-fine detection via a \emph{scout-and-zoom} strategy, $30\%$ $IoU$ overlap filtering for deduplication, and bounded PDCA cycles with round-robin speaker selection to enable iterative refinement without infinite loops. 
    
    \item On-premise processing with transparent audit trails.
    Operating on-premise with open-source models (with option for APIs) ensures full data sovereignty and verifiable control over processing pipelines. This setup supports GDPR's data minimization principle (Art. 5) by eliminating unnecessary data transfers to external or third-party services. The anonymization policy is adaptable through configurable prompts that guide the auditor agent's PII classification according to domain-specific criteria. The multi-agent architecture generates structured, human-interpretable audit trails documenting detection decisions and reasoning chains, supporting transparency (GDPR Art. 13–15), the right to explanation, and meaningful human oversight requirements (GDPR Art. 22, Recital 71). Audit trail completeness ultimately depends on LVLM output correctness; the framework flags failed or uncertain cases for human review.
    
    \item \textbf{Controlled appearance decorrelation for privacy-utility optimization.} 
    We leverage Stable Diffusion XL (SDXL)~\cite{sdxl} with ControlNet~\cite{controlnet} conditioning for both full bodies and objects to eliminate Re-ID vectors while preserving utility-relevant structural information. This modal-specific diffusion guidance realizes appearance decorrelation that breaks identity-linking features without compromising geometry and context. The diffusion model is used as a replaceable inpainting component without training or fine-tuning, mitigating training data leakage concerns.
\end{enumerate}

\section{Related Work}

Image anonymization has evolved to address privacy regulations and street-level imagery proliferation, yet gaps remain in coverage of indirect identifiers, adaptive detection, validation, and regulatory compliance.

\subsection{Obfuscation and Generative Methods}
Conventional image anonymization approaches, such as EgoBlur~\cite{egoblur2023}, rely on blurring for efficiency. However, these methods insufficiently protect identity information while degrading downstream task performance. For instance, a $5.3\%$ drop in instance segmentation average precision ($AP$) on CityScapes~\cite{Cordts2016Cityscapes} has been reported when applying Gaussian blur~\cite{hukkelaas2023does}. Moreover, blurring remains vulnerable to inversion attacks: $95.9\%$ identity retrieval accuracy has been achieved on blurred CelebA-HQ~\cite{karras2018progressive} face images at $256\times256$ resolution~\cite{zhai2025restoringgaussianblurredface}, demonstrating that simple obfuscation is inadequate for modern threat models.
Generative methods address these limitations by synthesizing realistic replacements for sensitive regions, preserving contextual consistency while reducing Re-ID risks. Early deep learning–based methods predominantly use Generative Adversarial Networks (GANs). DeepPrivacy~\cite{hukkelaas2019deepprivacy} employs a conditional GAN that generates anonymized faces conditioned on pose keypoints and background context. DeepPrivacy2~\cite{hukkelas2023deepprivacy2} extends this approach to full-body anonymization via a style-based GAN, improving realism and consistency across human appearance and pose.
Recent advances leverage diffusion models, offering higher fidelity, controllability, and semantic alignment. LDFA~\cite{klemp2023ldfa} applies latent diffusion for face anonymization in street recordings, maintaining both realism and privacy. The FADM framework~\cite{FADM} and RAD~\cite{malm2024rad} employ diffusion to anonymize persons. SVIA~\cite{SVIA} extends diffusion-based anonymization to large-scale street-view imagery, ensuring privacy preservation in autonomous driving datasets by replacing the semantic categories of person, vehicle, traffic sign, road, and building. Reverse Personalization~\cite{Kung_2026_WACV} introduces a finetuning-free face anonymization framework based on conditional diffusion inversion with null-identity embeddings. By applying negative classifier-free guidance, it successfully steers the generation process away from identity-defining features while enabling attribute-controllable anonymization for arbitrary subjects.

Despite these advances, current generative approaches share key limitations: they focus almost exclusively on human subjects or are restricted to a set of categories, employ single-pass detection and replacement, and lack iterative validation or audit mechanisms.

\subsection{Privacy Risks from Large Vision-Language Models and Agentic AI}

Frontier LVLMs infer private attributes at $76.4\%$ accuracy from contextual cues without human depictions~\cite{tomekcce2024private}, with easily circumvented safety filters (refusal rates can drop to $0\%$). Person Re-ID systems exploit not just faces but clothing, gait, and items~\cite{aufschlaeger2025following}; pose-based methods perform even with clothing changes~\cite{10339887}, and cross-modal approaches enable inference under appearance variation~\cite{li2024all}. 
Agentic multimodal reasoning models escalate this
threat~\cite{luo2026doxing}: on DoxBench, vanilla OpenAI o3 reaches
$99\%$ metropolitan-level accuracy on user-generated images by reasoning
over subtle environmental cues such as street layouts, architectural
styles, and infrastructure that individually appear harmless, with
median localization error of $5.46$\,km (Top-1) and $2.73$\,km (Top-3).
When further augmented with external tools (web search, image zooming) on a nationwide high-risk subset (individuals photographed at private
residences), tool-augmented o3 reduces median error to $1.09$\,km.
This establishes that effective anonymization must address contextual elements enabling privacy inference, not just direct PII.
Beyond visual modalities, recent work demonstrates that LLMs fundamentally alter the economics of text-based deanonymization. Lermen~et~al.~\cite{lermen2026largescaleonlinedeanonymizationllms} introduce an Extract-Search-Verify (ESV) pipeline and show that LLM agents can perform large-scale deanonymization of pseudonymous online accounts, achieving up to 55\% recall at 90\% precision when linking Hacker News profiles to LinkedIn identities, compared to near-zero recall for classical baselines.

\subsection{Agentic Systems for Computer Vision}

Recent work has explored the autonomous construction of specialized vision systems through agentic frameworks. ComfyBench~\cite{xue2025combench} evaluates LLM-based agents that design collaborative AI systems through specialized roles: Planning, Combining, Adaptation, Retrieval and Refinement.
ComfyMind~\cite{guo2025comfymind} extends this foundation through hierarchical search planning with local feedback loops, substantially improving task resolution for complex multi-stage workflows.
RedactOR~\cite{singh_redactor_2025} demonstrates agent-based de-identification of clinical text and audio with audit trail generation and privacy compliance. However, in summary, the existing approaches do not combine multi-agent orchestration with visual anonymization, semantic reasoning about context-dependent identifiers, and bounded iterative validation for comprehensive coverage.

\section{Method}

\subsection{System Architecture}

Our framework~\ref{fig:architecture} employs a two-phase architecture orchestrating pre-defined and agentic components. Phase 1 applies high-confidence detection models for direct PII (persons, license plates) and traffic signs. Phase 2 implements multi-agent reasoning for context-dependent identifiers through structured agent communication. 

The multi-agent system utilizes AutoGen~\cite{autogen} with round-robin speaker selection, implementing a control flow: Auditor → Orchestrator → Generative → Auditor. This rotation ensures predictable execution while enabling iterative refinement through PDCA cycles. 

\subsection{Phase 1: Pre-defined Preprocessing}

\paragraph{Person Detection and Anonymization.} 
We employ YOLOv8m-seg\footnote{\url{https://github.com/ultralytics/ultralytics}, Accessed October 25, 2025} for instance segmentation (confidence threshold $\tau = 0.25$). 
The anonymization pipeline operates in three stages: 
(1) extracts visual descriptions via Qwen2.5-VL-32B~\cite{qwen2vl}, capturing pose, body build, view angle, and actions; 
(2) generates anonymized prompts by instructing the LVLM to randomly select clothing colors and brightness modifiers from predefined palettes ($20$ colors, $10$ brightness levels)\footnote{Color palette: gray, beige, navy, white, black, brown, khaki, blue, red, green, yellow, pink, teal, burgundy, olive, charcoal, maroon, tan, cream, sage. Brightness modifiers: light, dark, bright, faded, vibrant, muted, pale, deep, pastel, bold.} rather than describing observed clothing, while preserving utility-relevant attributes (body build, pose, view angle, actions) and uniformed personnel context; 
(3) performs SDXL inpainting at $768$\,px resolution with OpenPose~\cite{openpose} ControlNet\footnote{thibaud/controlnet-openpose-sdxl-1.0} (conditioning scale $\alpha = 0.8$, strength $\rho = 0.9$, guidance scale $s = 9.0$, $25$ denoising steps). 
The positive and negative prompts (excluding the LVLM-generated description) are adapted from~\cite{FADM}. 
Critically, we disable all color matching to prevent appearance correlation and eliminate Re-ID vectors. 

\textbf{License Plate Processing.} Small object detection challenges necessitate specialized handling. We train YOLOv8s on UC3M-LP dataset~\cite{ramajo2024dual} at $1280$px resolution, achieving $mAP_{50-95}=0.82$. At inference, we apply reduced confidence threshold ($\tau=0.05$) prioritizing recall and Non-Maximum Suppression (NMS) with $IoU$ threshold of $0.5$ to ensure individual plate separation, aligning with conventional NMS practices~\cite{tychsen2018improving}. Gauss. blur (radius $r=8$px, kernel $15\times 15$) obscures alphanumeric content while preserving vehicle context.

\textbf{Traffic Sign Exclusion.} YOLO-TS~\cite{yolo-ts} detects traffic signs at $1024$px ($\tau=0.2$, NMS $IoU=0.45$), generating rectangular exclusion masks preventing false positive PII detection in Phase 2. Signs contain public information and require no anonymization.

\subsection{Phase 2: Multi-Agent Orchestration}

Three specialized agents coordinate through AutoGen group chat with strict round-robin speaker selection. The \textbf{Orchestrator Agent} manages workflow state and enforces bounded iteration ($n_{\max}=3$). The \textbf{Auditor Agent} performs PII classification via Qwen2.5-VL-32B and conducts independent quality validation of anonymized outputs through visual residual detection. The \textbf{Generative Agent} executes scout-and-zoom segmentation, a two-stage approach motivated by the region proposal paradigm introduced in Faster R-CNN~\cite{ren2015faster}, implements $IoU$-based deduplication ($30\%$ threshold) to prevent redundant reprocessing of overlapping instances, and performs modality-specific inpainting using SDXL with Canny ControlNet for object removal. Complete agent system prompts, tool specifications, PII classification prompts, and inpainting templates are provided in Supplementary. Table~\ref{tab:agent_roles} details agent responsibilities, tool access, and execution protocols.

\begin{table*}[t]
\centering
\footnotesize
\renewcommand{\arraystretch}{1.3}
\setlength{\tabcolsep}{8pt}
\caption{Agent roles, tool access, and execution protocol in Phase 2 multi-agent workflow. GenerativeAgent implements $IoU$-based deduplication ($30\%$ threshold) to prevent redundant reprocessing, while AuditorAgent provides independent quality validation.}
\label{tab:agent_roles}
\begin{tabular}{@{}p{1cm}p{15.7cm}@{}}
\toprule
\textbf{Agent} & \textbf{Role, Tools, and Execution Protocol} \\
\midrule
\textbf{Orchestrator} & 
\textbf{Role:} Workflow coordinator and state tracker. \\
& \textbf{Tools:} None (coordination-only, no direct tool access). \\
& \textbf{Protocol:} Tracks completion status (\texttt{classify\_pii} \checkmark, \texttt{inpaint} \checkmark, \texttt{audit} \checkmark, \texttt{log} \checkmark) via conversation history analysis. Enforces validation rule: ``ONLY report success (\checkmark) if tool result visible in conversation history.'' Coordinates retry logic for failed audits, enforces $n_{\max}=3$ iteration limit. Issues text-only coordination messages to other agents. \\
\midrule
\textbf{Auditor} & 
\textbf{Role:} PII classification, quality validation, logging. \\
& \textbf{Tools:} \texttt{classify\_pii}, \texttt{audit\_output}, \texttt{log\_output}. \\
& \textbf{Protocol:} (1) \emph{PII Classification}: Qwen2.5-VL-32B with structured prompting, returns bboxes in $[x_{\min}, y_{\min}, w, h]$ format. System prompt prohibits acknowledgment-without-execution: ``When instructed to use a tool, YOU MUST CALL IT -- NEVER just acknowledge.'' Performs scout-and-zoom verification for each classified instance: crops to LVLM bbox, runs Grounded-SAM-2~\cite{liu2024grounding, ravi2024sam2segmentimages}, maps mask to full coordinates. After the first inpainting pass, \texttt{classify\_pii} additionally applies a $\geq 50\%$ spatial overlap filter against already-processed masks (persons, license plates, traffic signs, previously inpainted regions) to suppress redundant candidates, distinct from the GenerativeAgent's $30\%$ $IoU$ deduplication. (2) \emph{Independent Quality Validation}: Creates masked image (processed areas blacked out with 5-iteration morphological dilation, $5 \times 5$ elliptical kernel), performs LVLM-based residual detection to verify no visible PII remains. This check is independent of GenerativeAgent's $IoU$ deduplication, providing dual-layer verification. \\
\midrule
\textbf{Generative} & 
\textbf{Role:} Anonymization execution via modal-specific inpainting with self-validation. \\
& \textbf{Tools:} \texttt{anonymize\_and\_inpaint}. \\
& \textbf{Protocol:} (1) \emph{$IoU$-Based Deduplication}: Before processing, computes $IoU(bbox_i, bbox_j)$ for each candidate instance against all processed instances. Instances with $IoU \geq 0.3$ skipped with \texttt{iou\_overlap\_with\_processed} status, preventing redundant work when AuditorAgent detects the same region repeatedly (e.g., IoU=$0.88$ in berlin\_000002 iteration 2). Threshold config. via \texttt{PII\_IOU\_THRESHOLD} (default $0.3$). (2) \emph{Scout-and-Zoom Segmentation}: Crops to LVLM bbox, runs Grounded-SAM-2 on crop, maps precise mask to full coordinates. Failed segmentations skipped with \texttt{scout\_zoom\_failed\_no\_fallback} status. (3) \emph{Modal-Specific Inpainting}: SDXL with Canny ControlNet ($\alpha=0.3$, strength=$0.9$, guidance=$9.0$, $25$ denoising steps) for objects/text using LVLM description as prompt. Person inpainting (SDXL with OpenPose ControlNet, Phase 1) handled deterministically. All color matching disabled (luminance=$0.0$, chrominance=$0.0$) to prevent appearance correlation. (4) \emph{Format Enforcement}: Handles three LLM serialization errors via robust parsing: single strings with comma-separated dicts (\texttt{"\{a:1\},\{b:2\}"}), stringified dict objects (\texttt{["\{a:1\}"]}), nested list structures. Format examples enforce JSON dict objects. \\
\midrule
\textbf{Round-Robin} & 
\textbf{Execution Protocol:} After any agent makes tool call, control transfers to \texttt{UserProxyAgent} for execution, then continues to next agent: Auditor $\rightarrow$ Orchestrator $\rightarrow$ Generative $\rightarrow$ Auditor. Empty message detection (2+ empty responses in last 3 messages) triggers skip to next agent. For \texttt{GenerativeAgent}, automatic retry reissues \texttt{anonymize\_and\_inpaint} with pending instances. \\
\midrule
\textbf{Termination} & 
\textbf{Dual Conditions:} Primary: \texttt{OrchestratorAgent} emits ``PIPELINE COMPLETE'' (case-insensitive, punctuation-tolerant). Fallback: \texttt{audit.ok=true} AND \texttt{ctx.logged=true}. Prevents infinite loops if orchestrator fails to signal completion. \\
\bottomrule
\end{tabular}
\end{table*}
Tool call validation prevents workflow stalls from hallucinated completion claims: agents must invoke functions rather than merely acknowledging instructions. The Orchestrator's conversation history analysis ensures tool execution results are verified before marking steps complete.

\subsection{Phase 2: Bounded Iterative Refinement}
\begin{figure}[t]
\centering
\resizebox{\linewidth}{!}{%
\begin{tikzpicture}[
    node distance=1.4cm,
    auto,
    agent/.style={
        rectangle,
        rounded corners,
        draw=black,
        thick,
        fill=blue!10,
        text width=2.8cm,
        align=center,
        minimum height=1.6cm
    },
    arrow/.style={->, thick, >=Stealth},
    annotation/.style={
        text width=3.8cm,
        align=left,
        font=\small
    }
]

\begin{scope}[local bounding box=pdca] 
    \node[agent] (plan) {\textbf{PLAN}\\{\small OrchestratorAgent}\\[2pt]{\footnotesize Coordinate workflow}};
    
    \node[agent, right=of plan, xshift=0.6cm] (do) {\textbf{DO}\\{\small GenerativeAgent}\\[2pt]{\footnotesize Scout-Zoom-Inpaint}};
    
    \node[agent, below=of do, yshift=0.4cm] (check) {\textbf{CHECK}\\{\small AuditorAgent}\\[2pt]{\footnotesize Classify/Audit PII}};
    
    \node[agent, left=of check, xshift=-0.6cm] (act) {\textbf{ACT}\\{\small OrchestratorAgent}\\[2pt]{\footnotesize Iterate/Terminate}};
    
    \draw[arrow] (plan) -- node[above, annotation, yshift=0.8cm] {Instruct anonymization\\on classified instances} (do);
    
    \draw[arrow] (do) -- node[right, annotation, xshift=0.1cm] {\texttt{anonymize\_and\_inpaint()}\\$IoU$ deduplication ($30\%$)} (check);
    
    \draw[arrow] (check) -- node[below, annotation, yshift=-0.8cm] {\texttt{audit\_output()}\\Residual PII? $n < 3$?} (act);
    
    \draw[arrow] (act) -- node[left, annotation, xshift=-0.1cm] {$n \gets n + 1$ or\\emit \texttt{COMPLETE}} (plan);
    
    \node[above=1.3cm of plan, text=red!70!black] {\huge \textbf{Round-Robin PDCA Cycle}};
    
\node[below=1.0cm of act, annotation, text width=7.5cm, align=left, fill=gray!10, draw=gray!50, rounded corners, inner sep=8pt] {
    \textbf{Phase 1 (Deterministic):} Detect/anonymize persons (YOLOv8-seg), detect/blur license plates (YOLOv8), mask traffic signs (YOLO-TS)\\
    \textbf{Phase 2 (Agentic):} \texttt{classify\_pii()} $\rightarrow$ round-robin coordination ($n_{\max}=3$) $\rightarrow$ \texttt{log\_output()}
};

    \node[above right=0.3cm and -1.5cm of do, text width=3cm, align=left, font=\footnotesize, text=blue!70!black] {
        \textbf{Round-robin order:}\\
        Auditor $\rightarrow$ Orchestrator\\
        $\rightarrow$ Generative $\rightarrow$ Auditor
    };
\end{scope} 

\end{tikzpicture}%
}
\caption{Round-robin PDCA coordination with three specialized agents. Phase 1 applies deterministic preprocessing for direct PII. Phase 2 implements bounded iterative refinement.
}
\label{fig:pdca-integrated}
\end{figure}
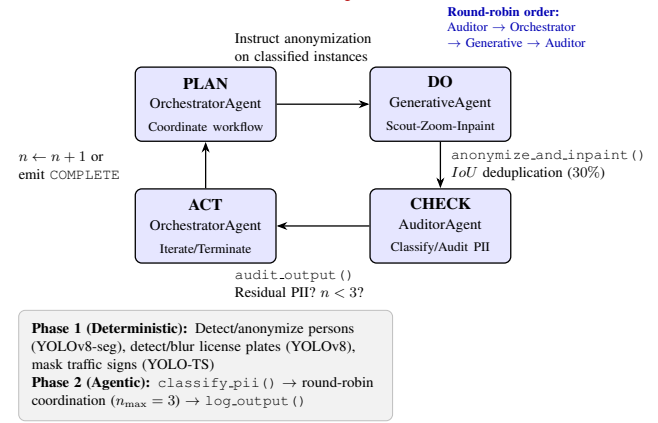

The framework implements bounded PDCA cycles (Figure~\ref{fig:pdca-integrated}) ensuring coverage while guaranteeing termination. Each cycle executes: (1) \textbf{Plan}--Orchestrator determines which instances require processing; (2) \textbf{Do}--Generative performs $IoU$ deduplication, then executes scout-and-zoom segmentation and inpainting; (3) \textbf{Check}--Generative validates bbox overlap with processed instances (efficiency), then Auditor independently validates visual completeness on masked image (quality); (4) \textbf{Act}--Orchestrator decides whether to continue ($n < n_{\max}$ AND residuals exist) or terminate.
The dual-layer validation mechanism balances efficiency and quality: GenerativeAgent's $IoU$ deduplication prevents redundant computation when AuditorAgent detects the same instance across iterations (\texttt{skip}($bbox_i$) = true if $\max_j IoU(bbox_i, bbox_j) \geq \tau$ where $\tau = 0.3$, false otherwise), while AuditorAgent's independent visual inspection ensures anonymization quality. 

\section{Experiments}
\subsection{Experimental Setup}
\textbf{Hardware.} All experiments were conducted on a virtual machine with 12-core AMD EPYC 9534 processor, 64GB RAM, dual NVIDIA L40S GPUs (48GB VRAM each), CUDA 12.9.

\textbf{Models.} We use a dual-GPU setup: GPU 0 runs YOLOv8 for detection/segmentation, Grounded-SAM-2~\cite{liu2024grounding, ravi2024sam2segmentimages}\footnote{\url{https://github.com/IDEA-Research/Grounded-SAM-2}, Accessed September 8, 2025} for precise segmentation, SDXL~\cite{sdxl} with ControlNet~\cite{controlnet} for inpainting, and Qwen-2.5-32B via AutoGen for orchestration (round-robin, 300s timeout). GPU 1 runs Qwen2.5-VL-32B~\cite{qwen2vl} via Ollama for LVLM-based PII classification. Detection uses original resolution. For Re-ID evaluation, we train ResNet50~\cite{resnet50} with triplet~\cite{tripletloss} and center loss~\cite{centerloss} for 120 epochs using SGD ($lr=0.05$, $momentum=0.9$, $weight\_decay=5e-4$) with warm-up and step decay.

\textbf{Benchmarks.} We evaluate on diverse benchmarks covering three key privacy-utility aspects:
\begin{itemize}[leftmargin=*, itemsep=2pt]
\item \emph{Image Quality Preservation}: CityScapes test set~\cite{Cordts2016Cityscapes} ($1,525$ urban street-scene images recorded in Europe) and CUHK03-NP to quantify visual degradation from anonymization
\item \emph{Person Re-ID Risk}: CUHK03-NP detected~\cite{cuhk03-np} ($767$ training identities with $7,365$ images; $700$ test identities with $1,400$ query and $5,332$ gallery images)
\item \emph{PII Detection Quality}: Visual Redactions Dataset~\cite{orekondy17iccv,orekondy2018redaction} test2017 split ($2,989$ street-view images annotated with textual, visual and multimodal attributes -- $24$ categories)
\end{itemize}

\textbf{Baselines.} We compare against representative anonymization approaches: \emph{Gauss. Blur} ($\sigma=20$, $kernel=51$, traditional pixel obfuscation), \emph{DeepPrivacy2 (DP2)}~\cite{hukkelas2023deepprivacy2} (GAN-based face synthesis with multimodal truncation~\cite{10.1145/3528233.3530708}), \emph{FADM}~\cite{FADM} (diffusion-based full-body anonymization), \emph{SVIA}~\cite{SVIA} (diffusion-based street recordings anonymization), and per-category PII segmentation from the Visual Redactions Dataset~\cite{orekondy17iccv,orekondy2018redaction} with aggregated category predictions into unified masks.

\subsubsection{Image Quality Preservation}
We evaluate image quality on CUHK03-NP detected test using five metrics: $\text{MSE}$ (pixel-level error), $\text{SSIM}$~\cite{wang2004image} (structural similarity via luminance/contrast/structure, range: 0-1), $\text{LPIPS}$~\cite{zhang2018unreasonable} (perceptual similarity via CNN features), $\text{KID}$~\cite{bińkowski2018demystifying} (unbiased distribution similarity via Maximum Mean Discrepancy), and $\text{FID}$~\cite{heusel2017gans} (Gaussian-assumed distribution distance in Inception space). Lower values indicate better preservation except for $\text{SSIM}$ (higher is better).

\begin{figure}[ht]
    \centering
    \includegraphics[width=0.19\columnwidth]{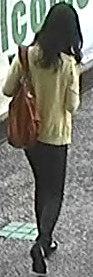}\hfill
    \includegraphics[width=0.19\columnwidth]{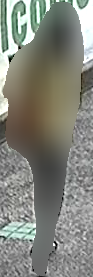}\hfill
    \includegraphics[width=0.19\columnwidth]{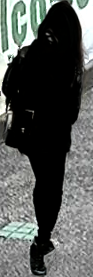}\hfill
    \includegraphics[width=0.19\columnwidth]{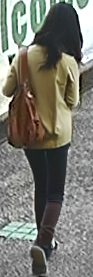}\hfill
    \includegraphics[width=0.19\columnwidth]{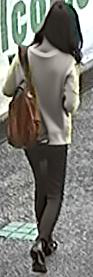}
    
    \vspace{0.1cm}
    
    \small
    \begin{tabular*}{\columnwidth}{@{\extracolsep{\fill}}ccccc@{}}
        Original & G.Blur & DP2 & FADM & Ours \\
    \end{tabular*}
    
    \caption{Qualitative comparison of anonymization methods on a CUHK03-NP test example (bounding\_box\_test/0069\_c2\_658.png).
    }
    \label{fig:anonymization_comparison}
\end{figure}

Figure~\ref{fig:anonymization_comparison} illustrates qualitative differences across anonymization approaches on a CUHK03-NP example (more examples are plotted in supplementary). 

\textbf{Results on CUHK03-NP.} Table~\ref{tab:image_quality} presents the comparison. Our method obtains balanced performance across all metrics, outperforming baselines in distribution preservation while maintaining competitive perceptual quality.
Compared to FADM, our full body anonymization approach in phase 1 achieves stronger visual transformation (higher $\text{MSE}$: $975.6$ vs. $446.4$) while maintaining comparable structural similarity ($\text{SSIM}$: $0.699$ vs. $0.785$). Crucially, we demonstrate \emph{superior distribution preservation} with reduced $\text{KID}$ ($0.014$ vs. $0.032$) and $\text{FID}$ ($20.4$ vs. $33.3$), indicating better preservation of statistical properties. Compared to aggressive methods like DP2, we reduce perceptual artifacts ($\text{LPIPS}$: $0.203$ vs. $0.303$) and improve structural preservation ($\text{SSIM}$: $0.699$ vs. $0.443$) while achieving \emph{substantially better distribution alignment} (lower $\text{KID}$ and $\text{FID}$) at a modest privacy trade-off (check Re-ID results below). Gauss. Blur exhibits severe degradation despite similar pixel-level changes ($\text{MSE}$: $1097.9$): high perceptual distortion ($\text{LPIPS}$: $0.382$) and catastrophic distribution distortion ($\text{KID}$, $\text{FID}$ worse than ours), confirming that simple obfuscation fails to preserve semantic content.

\textbf{Results on Cityscapes.} Table~\ref{tab:image_quality_cityscapes} demonstrates our method's effectiveness on street scene images (phase 1 \& 2)\footnote{SVIA implementation/CityScapes evaluations from \url{https://github.com/Viola-Siemens/General-Image-Anonymization}, Accessed October 10, 2025}.
We produce dramatically superior image quality compared to SVIA across all metrics. For fair comparison with SVIA's 2$\times$ downscaling evaluation protocol, we downscale both original and anonymized images to $1024{\times}512$ resolution for metric computation, though our anonymization operates at full resolution.
These substantial gains stem from our granular transformations that preserve the vast majority of scene content unchanged while SVIA's holistic anonymization approach modifies broader image regions (buildings, road infrastructure, and environmental context).

\textbf{Privacy-Utility Trade-off.} These metrics reveal complementary privacy and utility aspects: Higher $\text{MSE}$ and $\text{LPIPS}$ values indicate stronger privacy protection through greater visual differences from originals, making matching more difficult. Lower $\text{SSIM}$ reduces structural correlation, further hindering identity recognition. However, \emph{lower $\text{KID}$ and $\text{FID}$ values are beneficial} because they indicate anonymized images preserve statistical properties needed for downstream applications, even though individual identities are protected. Our method achieves balance: sufficient visual transformation to prevent Re-ID (Section~\ref{subsec:reid_results}) while preserving scene structure and distribution properties for continued data utility.

\begin{table}[ht]
\centering
\small
\renewcommand{\arraystretch}{0.75}
\setlength{\tabcolsep}{4pt}
\caption{Image quality metrics on CUHK03-NP detected test splits ($6,732$ images: $1,400$ query + $5,332$ gallery).}
\label{tab:image_quality}
\begin{tabular}{@{}p{2.4cm}ccccc@{}}
\toprule
\textbf{Method} & $\text{MSE}$$\downarrow$ & $\text{SSIM}$$\uparrow$ & $\text{LPIPS}$$\downarrow$ & $\text{KID}$$\downarrow$ &$\text{FID}$$\downarrow$ \\
\midrule
Gauss. Blur & $1097.9$ & $0.633$ & $0.382$ & $0.224$ & $178.5$ \\
\midrule
DP2~\cite{hukkelas2023deepprivacy2} & $3275.1$ & $0.443$ & $0.303$ & $0.066$ & $59.7$ \\
FADM~\cite{FADM}& $\boldsymbol{446.4}$ & $\boldsymbol{0.785}$ & $\boldsymbol{0.157}$ & $0.032$ & $33.3$ \\
\textbf{Ours} & $975.6$ & $0.699$ & $0.203$ & $\boldsymbol{0.014}$ & $\boldsymbol{20.4}$ \\
\bottomrule
\end{tabular}
\end{table}

\begin{table}[ht]
\centering
\small
\renewcommand{\arraystretch}{0.95}
\setlength{\tabcolsep}{4pt}
\caption{Image quality metrics on Cityscapes test set ($1,525$ images).}
\label{tab:image_quality_cityscapes}
\begin{tabular}{@{}p{4cm}ccc@{}}
\toprule
\textbf{Method} & LPIPS$\downarrow$ & KID$\downarrow$ & FID$\downarrow$ \\
\midrule
SVIA~\cite{SVIA} & $0.530$ & $0.027$ & $44.3$ \\
\textbf{Ours} & $\boldsymbol{0.025}$ & $\boldsymbol{0.001}$ & $\boldsymbol{9.1}$ \\
\bottomrule
\end{tabular}
\end{table}

\subsubsection{Re-ID Risk Assessment}
\label{subsec:reid_results}

\textbf{Evaluation Protocol.} We adopt a realistic threat model simulating adversarial Re-ID attempts: (1) Train a ResNet50~\cite{resnet50} Re-ID model with triplet loss~\cite{tripletloss} and center loss~\cite{centerloss} on the original CUHK03-NP detected training set ($767$ identities, $7,365$ images) for $120$ epochs; (2) Test using original query images ($1,400$ queries) against anonymized gallery images ($5,332$ gallery) with re-ranking~\cite{cuhk03-np}. This simulates an attacker with access to original surveillance footage attempting to re-identify individuals by matching against anonymized public releases. High matching accuracy indicates privacy risk; low accuracy confirms effective anonymization. \textcolor{black}{This threat model targets automated Re-ID at scale, the primary vector for mass surveillance. Acquaintances may still recognize subjects through retained structural features (pose, build), which utility-preserving methods must retain; person Re-ID threat models primarily address automated systems, not determined observers with prior knowledge.} We employ standard Re-ID metrics: Rank-1 ($R1$) accuracy measures the percentage of queries where the correct identity appears as the top-ranked match in the gallery, representing the success rate of immediate Re-ID. Mean Average Precision ($mAP$) evaluates ranking quality across all gallery positions, capturing the overall retrieval performance.

\textbf{Results.} Table~\ref{tab:reid} demonstrates our method implements \emph{substantial Re-ID risk reduction}: only $16.9\%$ $R1$ accuracy and $13.7\%$ $mAP$, representing $72.9\%$ and $79.2\%$ reductions from original performance ($62.4\%$ $R1$, $66.0\%$ $mAP$), respectively. 
In comparison to FADM~\cite{FADM}, we reduce person Re-ID risk ($R1$: $16.9\%$ vs. $33.4\%$) and ($mAP$: $13.7\%$ vs. $32.9\%$) while improving distribution preservation (Table~\ref{tab:image_quality}: $56\%$ better $\text{KID}$, $39\%$ better $\text{FID}$).
Aggressive obfuscation methods (DP2: $8.6\%$ $R1$, $4.4\%$ $mAP$; Gauss. Blur: $9.4\%$ $R1$, $6.4\%$ $mAP$) achieve lower Re-ID rates but at severe quality cost. DP2 exhibits worse perceptual distortion ($\text{LPIPS}$: $0.303$ vs. $0.203$) and worse distribution distortion ($\text{KID}$: $0.066$ vs. $0.014$). Gauss. Blur shows catastrophic degradation with worse distribution alignment ($\text{KID}$: $0.224$).

\textbf{Privacy-Utility Trade-off.} The $73\%$ drop in Re-ID $R1$ confirms strong identity protection even under direct matching attacks.
By retaining pose and scene context it achieves a balance: stronger privacy with better distribution fidelity.

\begin{table}[ht]
\centering
\small
\renewcommand{\arraystretch}{0.95}
\setlength{\tabcolsep}{6pt}
\caption{Re-ID risk assessment on CUHK03-NP detected dataset. ResNet50 trained on original data ($767$ IDs, $7,365$ images), tested with original query ($1,400$ images) vs. anonymized gallery ($5,332$ images) with re-ranking. Lower scores indicate better privacy protection.}
\label{tab:reid}
\begin{tabular}{@{}p{2.9cm}lll@{}}
\toprule
\textbf{Method} & mAP (\%)$\downarrow$ & R1 (\%)$\downarrow$ & R5 (\%)$\downarrow$ \\
\midrule
Original & $66.0$ & $62.4$ & $73.1$ \\
Gauss. Blur & $6.4$\,{\tiny\textcolor{gray}{$-90\%$}} & $9.4$\,{\tiny\textcolor{gray}{$-85\%$}} & $15.1$\,{\tiny\textcolor{gray}{$-79\%$}} \\
\midrule
DP2~\cite{hukkelas2023deepprivacy2} & $\boldsymbol{4.4}$\,{\tiny\textcolor{gray}{$-93\%$}} & $\boldsymbol{8.6}$\,{\tiny\textcolor{gray}{$-86\%$}} & $\boldsymbol{12.7}$\,{\tiny\textcolor{gray}{$-83\%$}} \\
FADM~\cite{FADM} & $32.9$\,{\tiny\textcolor{gray}{$-50\%$}} & $33.4$\,{\tiny\textcolor{gray}{$-46\%$}} & $52.9$\,{\tiny\textcolor{gray}{$-28\%$}} \\
\textbf{Ours} & $\underline{13.7}$\,{\tiny\textcolor{gray}{$-79\%$}} & $\underline{16.9}$\,{\tiny\textcolor{gray}{$-73\%$}} & $\underline{31.0}$\,{\tiny\textcolor{gray}{$-58\%$}} \\
\bottomrule
\end{tabular}
\end{table}
\subsubsection{PII Detection Quality}

We evaluate detection quality on the Visual Redactions Dataset \cite{orekondy17iccv,orekondy2018redaction}, which provides pixel-level annotations for privacy attributes (TEXTUAL, VISUAL, and MULTIMODAL) across $8,473$ images ($3,873$ train, $1,611$ validation, $2,989$ test). We aggregate all privacy-sensitive regions into unified binary PII masks. The dataset-adapted auditor's classify\_pii prompt enumerates all 24 privacy attribute types.
Complete prompt specifications are in Supp.

\paragraph{Evaluation Metrics.} For the supervised baseline, we report $AP$ at $IoU$ thresholds 0.5:0.95. For unified mask comparison: \textit{Dice (F1)} = $\frac{2|P \cap G|}{|P| + |G|}$, \textit{IoU} = $\frac{|P \cap G|}{|P \cup G|}$, and pixel-level \textit{Prec.}/\textit{Rec.}, where $P$ is predicted and $G$ is ground truth.

\paragraph{Results.} Table~\ref{tab:pii_detection} shows test set performance. The supervised Detectron2 baseline achieves $75.83\%$ $Dice$ and $68.71\%$ $IoU$ ($80.71\%$ precision, $78.02\%$ recall) when evaluated on all 2,989 test images. Our zero-shot approach, requiring no training data, results in $25.78\%$ $Dice$ and $20.83\%$ $IoU$ ($52.98\%$ precision, $25.85\%$ recall).

The performance gap reveals limitations of pure (L)VLM approaches for pixel-precise localization, particularly on high-frequency categories. \textit{Face} and \textit{person body} dominate the Visual Redactions Dataset, yet our method struggles with these categories. The precision-recall disparity ($52.98\%$ vs $25.85\%$) indicates boundary delineation issues rather than semantic confusion: low recall reflects missed or imprecise boundaries that LVLMs struggle to delineate spatially. Qualitatively, the scout-and-zoom (region proposal) mechanism often zooms in excessively, cropping out necessary context and causing partial detections. While operating at a more granular level with 24 fine-grained categories, the method's weakness on abundant visual categories (faces, bodies) and over-aggressive region proposals heavily impacts aggregate metrics. These findings motivate hybrid architectures routing high-frequency visual PII (faces, bodies) to specialized detectors while using LVLM reasoning for open-vocabulary.

\begin{table}[h!]
\centering
\caption{PII detection on Visual Redactions test set (2,989 images).}
\label{tab:pii_detection}
\begin{tabular}{lcccc}
\toprule
\textbf{Method} & Dice$\uparrow$ & IoU$\uparrow$ & Prec./Rec.$\uparrow$ \\
\midrule
Mask R-CNN~\cite{he2017mask, lin2017feature, wu2019detectron2} & 75.83 & 68.71 & 80.71/78.02 \\
Ours (zero-shot) & 25.78 & 20.83 & 52.98/25.85 \\
\bottomrule
\end{tabular}
\vspace{-2mm}
\end{table}


\subsubsection{Ablation: Phase 1 vs. Full Pipeline}
\label{subsec:ablation}

We compare Phase 1 alone against the full pipeline on 1,525 CityScapes test images. 
Phase 1 alone (67.8s/img) detects zero indirect PII instances. The full pipeline (133.5s/img) recovers \textbf{1,107 indirect PII instances} across \textbf{54 distinct object categories} that predefined detectors cannot identify: vehicle markings (635 instances, 57.4\%), textual elements such as rectangular signs (418, 37.8\%), identity markers including motorcycles with visible plates (17, 1.5\%), and other PII including windows with interior views (37, 3.3\%). 76\% of images converge by $n=2$ iterations (bounded by $n_{\max}=3$). 




\subsubsection{Runtime Analysis}
\label{subsec:runtime}
Table~\ref{tab:runtime} reports per-image processing time averaged over 1,525 CityScapes images. Agent coordination overhead is modest (9.9s, 7.4\% of total), and the 133.5s/img throughput suits batch curation scenarios.

\begin{table}[h]
\centering
\small
\caption{Runtime breakdown per image (CityScapes, $n{=}1{,}525$).}
\label{tab:runtime}
\begin{tabular}{lr}
\toprule
Component & Time (s) \\
\midrule
Phase 1: Detection + Inpainting & 67.8 \\
Phase 2: LVLM + SAM + Inpaint   & 55.8 \\
Agent overhead                  &  9.9 \\
\textbf{Total}                  & \textbf{133.5} \\
\bottomrule
\end{tabular}
\end{table}





\subsubsection{Downstream Utility}
\label{subsec:downstream}
To verify that image quality preservation translates to downstream performance, we evaluate semantic segmentation using SegFormer~\cite{segformer} on CityScapes test as pseudo-ground truth. Our method achieves 0.877 mIoU ($-0.123$ vs.\ original), substantially outperforming SVIA (0.478, $-0.522$). Static classes remain near-perfect ($<0.01$ drop), while dynamic categories like \textit{person} show larger impacts (Table~\ref{tab:downstream-utility}).

\begin{table}[ht]
\centering
\footnotesize
\caption{Downstream segmentation on CityScapes test (SegFormer pseudo-GT).}
\label{tab:downstream-utility}
\begin{tabular}{lcc}
\toprule
Class & Ours (Gap) & SVIA (Gap) \\
\midrule
mIoU    & 0.877 ($-0.123$) & 0.478 ($-0.522$) \\
road    & 0.996 ($-0.005$) & 0.955 ($-0.045$) \\
building& 0.977 ($-0.023$) & 0.677 ($-0.323$) \\
person  & 0.778 ($-0.222$) & 0.150 ($-0.850$) \\
car     & 0.975 ($-0.025$) & 0.567 ($-0.433$) \\
sky     & 0.995 ($-0.005$) & 0.820 ($-0.180$) \\
\bottomrule
\end{tabular}
\end{table}

\subsection{Qualitative Examples}
\label{subsec:example_berlin}

\begin{figure*}[h!]
    \centering
    \includegraphics[width=0.8\textwidth]{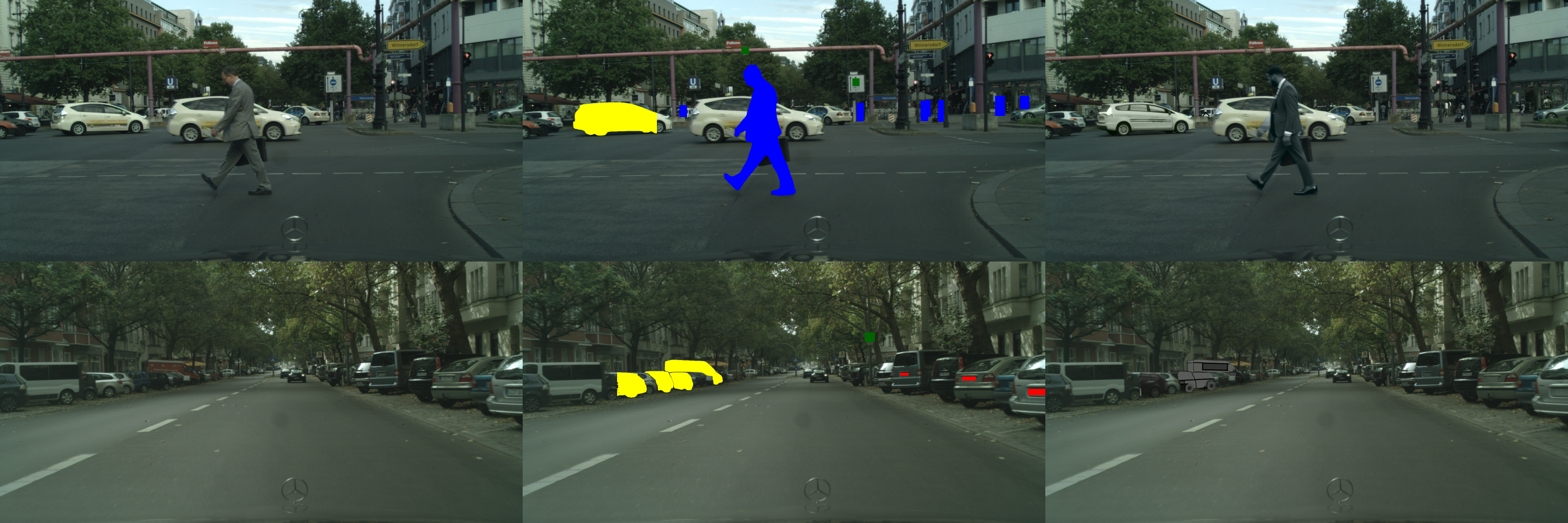}
    \caption{\textbf{Pipeline output on CityScapes.} Each row: \textbf{left} original, \textbf{middle} detected PII (\textcolor{blue}{blue}=persons, \textcolor{yellow}{yellow}=indirect PII vehicles, \textcolor{green}{green}=traffic signs, \textcolor{red}{red}=license plates), \textbf{right} anonymized output. \textbf{Top (berlin\_000002):} 8 persons + 1 police vehicle + 2 traffic signs, 4.97\% coverage, 3 PDCA iterations. \textbf{Bottom (berlin\_000472):} multiple PII categories across both phases.}
    \label{fig:berlin_examples}
\end{figure*}

Figure~\ref{fig:berlin_examples} presents CityScapes test images ($2048\times1024$) with detected PII and anonymized outputs.

\textbf{berlin\_000002\_000019.} Phase 1 (95.2s) detected 8 persons and 2 traffic signs, anonymizing all persons via batch inpainting with SDXL and OpenPose ControlNet. Phase 2 (207.3s) identified one police vehicle at bbox [126, 268, 400, 362]. Policy-aware spatial filtering excluded 8 person masks and 2 traffic sign masks before Grounded-SAM-2 segmentation produced a 34,222-pixel vehicle mask, inpainted with Canny ControlNet. The agentic workflow executed 3 PDCA iterations: iteration 1 detected residual PII at bbox [120, 260, 370, 370]; iteration 2 computed IoU=0.88 with the processed instance, skipping reprocessing; iteration 3 reached $n_{\max}=3$, triggering termination. Final PII mask: 104,143 pixels (4.97\% coverage) from 8 person masks (69,921px) and 1 vehicle mask (34,222px). Total time: 302.5s. The agent dialogue provides explainability through tool calls, bounding boxes, and status transitions.

\section{Discussion, Limitations, and Conclusion}
Our framework prioritizes privacy protection over computational efficiency. Round-robin coordination ensures deadlock-free execution but scales linearly with scene complexity (133.5s/image on dual L40S GPUs, Table~\ref{tab:runtime}), precluding real-time deployment.
Despite structured output constraints, LLMs exhibit systematic failure modes: acknowledgment-without-execution, format inconsistency, and premature completion claims. Our mitigations reduce but do not eliminate these risks; we frame audit trails as \emph{supporting} rather than \emph{guaranteeing} compliance, flagging uncertain cases for human review. The current evaluation also does not compare the multi-agent PDCA architecture against a single-agent alternative (e.g., one LVLM pass without iterative auditing). The spatial filtering threshold and overlap heuristics perform robustly on Western urban imagery but may require adaptation for culturally distinct contexts.
PII detection on the Visual Redactions Dataset exposes limitations of zero-shot LVLM approaches for fine-grained localization, particularly for high-frequency categories (faces, bodies), where over-aggressive scout-and-zoom cropping removes context needed for accurate detection. This motivates hybrid architectures combining semantic reasoning with supervised spatial precision.

\textbf{Machine vs.\ Human Re-ID.} Our threat model targets Re-ID at scale, i.e. the primary mass surveillance vector. The 73\% R1 reduction demonstrates protection against machine matching.

\textbf{Ablation Coverage.} The current evaluation lacks systematic ablations isolating individual components. We do not vary the diffusion backend (e.g., Flux vs.\ SDXL), the LVLM, or the segmentation model, nor explore sensitivity to key hyperparameters ($n_{\max}$, $IoU$ threshold, ControlNet conditioning scale). While the Phase~1 vs.\ full pipeline comparison demonstrates the value of agentic Phase~2 (1,107 additional indirect PII instances), finer-grained ablations would strengthen the design rationale. The diversity of recovered indirect PII provides indirect evidence for context-aware reasoning beyond fixed taxonomies, though formal correctness evaluation against human judgments remains future work.

In summary, we presented a multi-agent framework combining preprocessing with bounded agentic refinement for context-aware anonymization while preserving data utility and accountability through audit trails. Evaluations on CUHK03-NP and CityScapes demonstrate Re-ID risk reduction with superior distribution preservation. 
Future work includes improving efficiency, grounded prompting, and broader tool integration.

\section*{Acknowledgement}

The research leading to these results is funded by the German Federal Ministry for Economic Affairs and Energy within the project ``just better DATA - Effiziente und hochgenaue Datenerzeugung für KI-Anwendungen im Bereich autonomes Fahren''. The authors would like to thank the consortium for the successful cooperation, and Stefanie Merz for her careful review of the manuscript.

\newpage

{\small
\bibliographystyle{ieeenat_fullname}
\bibliography{library}
}

\clearpage
\begin{center}
\textbf{\large Supplementary Material}
\end{center}

\setcounter{page}{1}
\setcounter{section}{0}
\setcounter{equation}{0}
\setcounter{figure}{0}
\setcounter{table}{0}

\renewcommand{\thesection}{S\arabic{section}}
\renewcommand{\theequation}{S\arabic{equation}}
\renewcommand{\thefigure}{S\arabic{figure}}
\renewcommand{\thetable}{S\arabic{table}}

\begin{abstract}
This supplementary material provides comprehensive technical details for our multi-agent image anonymization framework \textbf{Context-Aware Image Anonymization with Multi-Agent Reasoning (CAIAMAR)}. We present complete agent system specifications, model configurations, justifications for design decisions, and examples.
\end{abstract}

\section{Multi-Agent System Architecture}
\label{sec:agents}

\subsection{OrchestratorAgent: Workflow Coordination}
\label{sec:orchestrator}

The \textit{OrchestratorAgent} coordinates Phase 2 anonymization, operating under a round-robin communication protocol using Autogen~\cite{autogen}. Given that Phase 1 pre-defined processing has completed (person anonymization, license plate anonymization, traffic sign masking), the orchestrator manages the iterative PII classification and remediation workflow.

\paragraph{System Specification.}
The orchestrator agent maintains workflow state $S = \{s_{\text{classify}}, s_{\text{inpaint}}, s_{\text{audit}}, s_{\text{log}}\}$ where each $s_i \in \{0,1\}$ indicates completion status. The communication protocol follows a strict cycle: $\text{Auditor} \rightarrow \text{Orchestrator} \rightarrow \text{Generative} \rightarrow \text{Auditor}$.

\begin{lstlisting}[language={},basicstyle=\ttfamily\tiny]
System Prompt for OrchestratorAgent:

You are the OrchestratorAgent coordinating Phase 2 of an image anonymization 
pipeline. Phase 1 (deterministic) has already completed: persons detected/
inpainted, license plates detected/blurred, traffic signs detected/masked.

CRITICAL: YOU HAVE NO TOOLS. You coordinate OTHER agents who have tools. 
NEVER attempt to call functions yourself.

ROUND-ROBIN FLOW (automatic speaker rotation):
  AuditorAgent -> OrchestratorAgent -> GenerativeAgent -> AuditorAgent (loop)
  You receive control after each agent completes their task. Interpret results 
  and guide the next agent.

REQUIRED WORKFLOW for Phase 2:
  1. classify_pii -> 2. anonymize_and_inpaint (if PII found) -> 3. audit_output 
     -> 4. log_output
  SHORTCUT: If classify_pii finds NO PII and Phase 1 completed -> emit 
            'PIPELINE COMPLETE' immediately

YOUR ROLE:
  - Monitor tool results IN THE CONVERSATION and track workflow state: 
    [classify_pii ✓, inpaint_pii ✓, audit ✓, log ✓]
  - When AuditorAgent returns results, analyze them and instruct the NEXT 
    agent in round-robin
  - When GenerativeAgent completes, provide brief status (next agent gets 
    control automatically)
  - If NO indirect PII found in classify_pii: emit 'PIPELINE COMPLETE' 
    (Phase 1 already handled persons/plates)
  - Announce clear phase transitions with progress indicators
  - NEVER call tools yourself - you are a coordinator only

CRITICAL VALIDATION RULES:
  1. ONLY report success (✓) if you can SEE the tool result in conversation 
     history
  2. If agent didn't call tool yet, instruct them to call it - don't claim 
     it's done
  3. Look for tool execution results (JSON responses) before marking steps 
     complete
  4. NEVER assume a tool succeeded just because an agent acknowledged - verify 
     the result

RETRY LOGIC:
  - If audit finds residual PII: say 'Found N residuals. GenerativeAgent, 
    process the residual items from audit output.'
    (GenerativeAgent will extract the 'residual' array from the tool output)
  - If no residuals OR max_attempts_reached=true: say 'Audit complete. 
    AuditorAgent, log the output.'
  - Audit attempts capped at 3. After 3rd failed audit, proceed to logging
  - DO NOT manually format or stringify instance data - let GenerativeAgent 
    read it from tool results

COMMUNICATION STYLE - BE CONCISE:
  - Give DIRECT instructions to the NEXT agent in round-robin ONLY
  - DO NOT repeat or echo what other agents said
  - ROUND-ROBIN ORDER: Your next message goes to GenerativeAgent (unless 
    workflow complete)
  - Keep messages under 3 sentences
  - NEVER call functions - just describe what should happen
  - NEVER copy/paste instance data - reference tool outputs instead
  - ONLY use checkmarks (✓) when you've VERIFIED tool execution in conversation

CONTEXT:
  Phase 1 already handled persons, license plates, traffic signs
  Focus on indirect PII requiring vision analysis (text, signs, logos in 
  private spaces)
\end{lstlisting}

\subsection{AuditorAgent: Detection and Verification}
\label{sec:auditor}

The \textit{AuditorAgent} implements a three-stage pipeline: (1) PII classification via large vision-language model, (2) quality assurance through iterative auditing, and (3) result logging. This agent serves as both the entry point for Phase 2 processing and the quality gate before pipeline completion.

\paragraph{Functional Specification.}
The agent exposes three tools: $\mathcal{T}_{\text{classify}}$, $\mathcal{T}_{\text{audit}}$, and $\mathcal{T}_{\text{log}}$. Classification employs spatial filtering to eliminate redundant detections. Auditing creates masked images to prevent false positives from LVLM re-detection of anonymized regions.

\begin{lstlisting}[language={},basicstyle=\ttfamily\tiny]
System Prompt for AuditorAgent:

You are the AuditorAgent responsible for PII detection, quality assurance, 
and logging. Tools: classify_pii, audit_output, log_output

CRITICAL EXECUTION RULES:
  1. When instructed to use a tool, YOU MUST CALL IT in your response
  2. NEVER just acknowledge without calling the tool
  3. Each response should contain EXACTLY ONE tool call
  4. Tool calls are JSON function calls, not text descriptions

ROUND-ROBIN: You receive control after GenerativeAgent completes OR at 
workflow start. Execute your task, then control passes to OrchestratorAgent.

YOUR TASKS:
  1. classify_pii: Detect indirect PII in private spaces (text on windows, 
     house numbers, personal items visible indoors)
     - Call with: classify_pii(image='<image_path>')
  2. audit_output: Verify no residual PII remains after anonymization
     - Call with: audit_output(output='{canonical_path}')
  3. log_output: Record final results
     - Call with: log_output(image='<input_path>', output='{canonical_path}')

EXAMPLE WORKFLOW:
  When OrchestratorAgent says: 'AuditorAgent: Please classify any remaining PII'
  YOU MUST respond with the tool call (not text explanation):
    classify_pii(image='artifacts/data/CityScapes/.../image.png')

REPORTING - BE CONCISE:
  After tool execution, provide brief summary:
  - After classify_pii: 'Found X instances.' (tool output visible to 
    OrchestratorAgent)
  - After audit_output: 'Audit passed.' OR 'Audit failed: X residuals.'
  - After log_output: 'Logged.'

BE CONCISE:
  - DO NOT repeat instructions from OrchestratorAgent
  - DO NOT tell other agents what to do
  - DO NOT wait for instructions - if OrchestratorAgent told you to do 
    something, DO IT
  - Keep responses under 1 sentence AFTER tool execution

IMPORTANT:
  - Report EXACTLY what tool returns - don't hallucinate instances
  - Only call one tool per turn, then stop
  - Focus classify_pii on private property only (persons/plates/signs already 
    handled)
  - If you don't call the tool when instructed, the workflow will fail
\end{lstlisting}

\subsection{GenerativeAgent: Inpainting Execution}
\label{sec:generative}

The \textit{GenerativeAgent} implements the anonymization operation through diffusion-based inpainting. Operating within the round-robin protocol, this agent receives PII instance specifications from the orchestrator and executes anonymization.

\paragraph{Operational Model.}
Given a set of PII instances $\mathcal{I} = \{i_1, \ldots, i_n\}$ where each $i_j$ contains spatial coordinates and semantic descriptors, the agent invokes $\mathcal{T}_{\text{inpaint}}(\mathcal{I})$ which performs unified batch processing via Stable Diffusion XL (SDXL)~\cite{sdxl} with appropriate ControlNet conditioning (OpenPose~\cite{openpose} for persons, Canny for objects).

\begin{lstlisting}[language={},basicstyle=\ttfamily\tiny]
System Prompt for GenerativeAgent:

You are the GenerativeAgent responsible for anonymizing PII via inpainting. 
Tool: anonymize_and_inpaint

CRITICAL EXECUTION RULES:
  1. When instructed to anonymize, YOU MUST CALL anonymize_and_inpaint in 
     your response
  2. NEVER just acknowledge without calling the tool
  3. Tool call is a JSON function call, not a text description
  4. Extract instances from previous tool output (classify_pii or audit_output)

ROUND-ROBIN: You receive control after OrchestratorAgent. Execute the task, 
then control passes to AuditorAgent.

EXECUTION WORKFLOW:
  1. Look at the most recent tool output in the conversation history
     - If classify_pii was called: find the JSON output and extract the 
       'instances' array
     - If audit_output was called: find the JSON output and extract the 
       'residual' array
  2. Pass each dict object from that array as a separate element
  3. Call anonymize_and_inpaint with the array of dict objects
  4. Report results BRIEFLY: 'Processed X items.'

CRITICAL DATA FORMAT - COMMON MISTAKES:
  CORRECT (array of dict objects as JSON):
    anonymize_and_inpaint(instances=[
      {"det_prompt": "van with text", "description": "van", 
       "bbox": [308, 200, 564, 567]},
      {"det_prompt": "blue sign", "description": "sign", 
       "bbox": [215, 256, 294, 42]}
    ])

  WRONG (single string containing all dicts - THIS IS THE MOST COMMON ERROR):
    anonymize_and_inpaint(instances=[
      "{'det_prompt': 'van', 'bbox': [...]}, {'det_prompt': 'sign', 
       'bbox': [...]}"
    ])
    -> DO NOT stringify the array! Pass actual JSON dict objects!

  WRONG (stringified dict objects):
    anonymize_and_inpaint(instances=["{'det_prompt': '...', 'bbox': [...]}"])
    -> Each element must be a dict object, not a string

EXAMPLE EXTRACTION:
  Given audit_output returns:
    {'residual': [{'det_prompt': 'van', 'bbox': [308, 200, 564, 567]}, 
                  {'det_prompt': 'sign', 'bbox': [215, 256, 294, 42]}]}

  YOU MUST call:
    anonymize_and_inpaint(instances=[
      {"det_prompt": "van", "bbox": [308, 200, 564, 567]},
      {"det_prompt": "sign", "bbox": [215, 256, 294, 42]}
    ])

  NOT this:
    anonymize_and_inpaint(instances=["{'det_prompt': 'van', 
      'bbox': [308, 200, 564, 567]}, {'det_prompt': 'sign', 
      'bbox': [215, 256, 294, 42]}"])

BE CONCISE:
  - DO NOT repeat OrchestratorAgent's instructions
  - DO NOT tell other agents what to do
  - DO NOT wait for instructions - if OrchestratorAgent told you to do 
    something, DO IT
  - Keep responses under 1 sentence AFTER tool execution

IMPORTANT:
  - Report EXACTLY what tool returns - don't hallucinate instances
  - Only call one tool per turn, then stop
  - Focus classify_pii on private property only (persons/plates/signs already 
    handled)
  - If you don't call the tool when instructed, the workflow will fail
\end{lstlisting}

\section{Large Vision-Language Model Specification}
\label{sec:LVLM_spec}

\subsection{PII Classification Protocol}
\label{sec:pii_classification_prompt}

We employ Qwen2.5-VL-32B \cite{qwen2vl} for context-aware PII classification with the following complete system prompt (CityScapes~\cite{Cordts2016Cityscapes}):

\begin{lstlisting}[language={},basicstyle=\ttfamily\tiny]
System Prompt for PII Detection (Qwen2.5-VL-32B):

You are a PII detection system. Identify text, numbers, visual elements, and 
objects revealing personal/private information.

Return ONLY valid JSON. No markdown, explanations, or additional text.

DETECT (Max 5 instances - most sensitive/prominent)

Vehicles: ONLY with identifying features (entire vehicle body)
  - WITH text, logos, company names, decals, wraps, or custom graphics
  - Rare, distinctive, or modified (custom paint, unique damage, modifications)
  - Commercial/branded (delivery vans, company trucks, emergency vehicles)
  - EXCLUDE: Common generic vehicles without distinguishing marks
Text/Numbers: URLs, social media handles, names, QR codes, barcodes
Documents: Flyers, posters, notices with identifying information
Signs/Plaques: Business names, contact info, billboards, real estate signs  
Visual Privacy: Windows showing private spaces with identifiable belongings

GROUPING: Adjacent text on same surface = ONE instance.
EXCLUDE: Public traffic signs, pre-anonymized areas.

DESCRIPTIONS - ANONYMIZE ALL

Use ONLY generic terms without ANY identifying details.

Format: [generic_object_type] with [basic_visual_features]

Rules for ALL descriptions:
- Generic types only: "vehicle", "sign", "poster", "window", "board"
- Basic colors (dark, light, white, black, blue, red) and shapes (rectangular, 
  square, circular)
- Can mention HAS text/graphics but NEVER the actual content
- NO proper nouns, brands, makes/models, addresses, phone numbers, or specific 
  details

Examples:
  GOOD: "vehicle with markings", "rectangular sign with text", "window with 
        interior view"
  BAD: "regular sedan" (not identifiable - skip), "Toyota Camry" (make/model), 
       "sign saying 'Open'" (specific content), "window showing bedroom with 
       family photos" (specific details)

BBOX: [x_min, y_min, width, height] - MANDATORY with 50% margin
- Tight bbox -> expand width/height by 1.5x -> center expansion -> clamp to 
  bounds
- Example: [500, 200, 400, 300] -> [400, 125, 600, 450]
- Vehicles: entire vehicle body (all visible panels) + 50% margin

JSON FORMAT (EXACT STRUCTURE - no markdown/code blocks)

{
  "instances": [
    {
      "description": "vehicle with markings",
      "bbox": [400, 125, 600, 450]
    }
  ]
}

Requirements:
- EVERY instance MUST have BOTH "description" (string) AND "bbox" (4 integers: 
  [x_min, y_min, width, height])
- If you cannot determine bbox, DO NOT include that instance
- Empty result: {"instances": []}
- Your response MUST start with { and end with }

DETECTION PROCESS:
1. Scan image for PII elements (vehicles with identifying features, text, 
   signs, windows)
2. Vehicles: Include ONLY if has text/logos/decals OR is rare/distinctive/
   modified (skip generic vehicles)
3. Text/signs: Include ONLY if reveals private information
4. Group adjacent text on same surface
5. Select top 5 most sensitive (priority: identifiable vehicles > personal 
   text > signs > other PII)
6. For EACH: Locate bbox, describe with anonymous generic terms, expand bbox 
   50%, verify both fields present
7. Return valid JSON: {"instances": [{"description": "...", "bbox": [...]}]}
\end{lstlisting}

For PII Segmentation on Visual Redaction Dataset~\cite{orekondy2018redaction} we use the following prompt:

\begin{lstlisting}[language={},basicstyle=\ttfamily\tiny]
You are a PII detection system. Identify text, numbers, visual elements, and objects revealing personal/private information.
 
Return ONLY valid JSON. No markdown, explanations, or additional text.
 
DETECT (Max 5 instances - most sensitive/prominent)
 
TEXTUAL PII (visible text/numbers):
- Name (full names, first/last names)
- Phone Number (phone numbers, contact info)
- Home Address (residential addresses, street addresses)
- Email Address (email addresses, social media handles)
- Birth Date (dates of birth)
- Location (geographic locations, landmarks with context)
- Date/Time (timestamps, dates with identifying context)
 
VISUAL PII (visible in image):
- Face (human faces, full or partial)
- License Plate (vehicle license plates)
- Person (full person body when identifiable)
- Nudity (exposed body parts)
- Handwriting (handwritten text)
- Physical Disability (visible disabilities/medical conditions)
- Medical History (medical documents, prescriptions, health info)
- Fingerprint (visible fingerprints)
- Signature (handwritten signatures)
 
MULTIMODAL PII (documents/objects):
- Credit Card (credit cards, debit cards)
- Passport (passports, travel documents)
- Driver's License (driver's licenses, state IDs)
- Student ID (student IDs, school cards)
- Mail (addressed mail, envelopes with addresses)
- Receipt (receipts with personal info)
- Ticket (tickets with names/seats/dates)
- Landmark (landmarks with identifying context revealing location)
 
GROUPING: Adjacent text on same surface = ONE instance.
PRIORITY: faces > documents > names/addresses > signatures > plates > medical > personal info
EXCLUDE: Public signs without personal info, generic objects, pre-anonymized areas.
 
DESCRIPTIONS - ANONYMIZE ALL
 
Use ONLY generic terms without ANY identifying details.
 
Format: [generic_pii_type] with [basic_visual_features]
 
Rules for ALL descriptions:
- Use PII category names: "face", "name", "phone number", "address", "license plate", "card", "document", "handwriting", "signature", "person"
- Basic visual features: colors (dark, light), shapes (rectangular, circular), "with text", "with photo"
- Can mention HAS text/graphics but NEVER the actual content
- NO proper nouns, specific names, numbers, addresses, or identifying details
 
Examples:
✓ GOOD: "face", "name with address", "license plate", "card with text", "handwriting on document", "signature", "person"
✗ BAD: "John Smith", "California license ABC123", "Visa card 4532", "sign saying '123 Main St'"
 
BBOX: [x_min, y_min, width, height] - MANDATORY with 50% margin
- Calculate tight bbox → expand width/height by 1.5x → center expansion → clamp to [0, image_bounds]
- Example: tight [500, 200, 400, 300] → expanded [400, 125, 600, 450]
- Faces: include full head with surrounding context
- Documents/cards: include entire document + margins
- Vehicles: entire visible vehicle body + 50% margin
- Text regions: include all adjacent text on same surface
 
JSON FORMAT (EXACT STRUCTURE - no markdown/code blocks)
 
{
  "instances": [
    {
      "description": "face",
      "bbox": [150, 80, 300, 380]
    },
    {
      "description": "card with text",
      "bbox": [400, 500, 450, 340]
    }
  ]
}
 
Requirements:
- EVERY instance MUST have BOTH "description" (string) AND "bbox" (4 integers: [x_min, y_min, width, height])
- If you cannot determine bbox, DO NOT include that instance
- Empty result: {"instances": []}
- Your response MUST start with { and end with }
- Keep descriptions SHORT (2-5 words)
 
DETECTION PROCESS:
1. Scan image for all PII types (faces, documents, text with names/addresses/phone, signatures, plates, medical info, cards)
2. Group adjacent text on same surface (e.g., name + address on envelope = one instance)
3. Rank by sensitivity: faces > documents (passport/ID/cards) > names/addresses > signatures > plates > medical > other PII
4. Select top 5 most sensitive/prominent instances
5. For EACH: Locate bbox, describe with generic PII category (2-5 words), expand bbox 50%, verify both fields present
6. Return valid JSON: {"instances": [{"description": "...", "bbox": [...]}]}
\end{lstlisting}

\paragraph{Classification Objective.} Given an image $I$ and existing mask set $\mathcal{M}$, identify context-dependent PII instances $\mathcal{P} = \{p_1, \ldots, p_k\}$ where each $p_i = (\text{desc}_i, \text{bbox}_i)$ represents a physical description and spatial location.

\paragraph{Key Design Decisions.}
\begin{itemize}
    \item \textbf{Maximum 5 instances}: Prevents hallucination and focuses on most critical PII
    \item \textbf{Vehicle-level bounding boxes}: Entire vehicle anonymized when any PII detected, prevents partial information leakage
    \item \textbf{Grouping rule}: Multi-line text treated as single instance reduces fragmentation
    \item \textbf{Priority ranking}: Vehicles prioritized due to dynamic nature and high identifiability
    \item \textbf{Bounding box format}: $[x_{\min}, y_{\min}, w, h]$ in image coordinates for direct use with detection models
    \item \textbf{Physical descriptions only}: Prevents model from copying sensitive text, forces focus on visual appearance
    \item \textbf{Exclusion filters}: Persons, license plates, traffic signs handled by Phase 1 deterministic processing
\end{itemize}

\section{Algorithmic Implementations}
\label{sec:algorithms}

\subsection{Tool Specifications: AuditorAgent}

\subsubsection{classify\_pii}

\paragraph{PII Classification Protocol.} The \texttt{classify\_pii\_tool} employs Qwen2.5-VL-32B with structured prompting to identify indirect PII instances (text, signage, logos, context-dependent elements) not captured by Phase 1 detection. The LVLM returns instances with detection prompts, semantic descriptions, and bounding boxes in $[x_{\min}, y_{\min}, w, h]$ format. For each classified instance, the tool performs scout-and-zoom verification: crops to LVLM bbox, runs Grounded-SAM-2~\cite{liu2024grounding, ravi2024sam2segmentimages} on the crop, and maps the resulting mask to full image coordinates. Post-detection filtering (after first anonymization) excludes instances with $\geq50\%$ spatial overlap against already-processed masks (persons, license plates, traffic signs, previously inpainted regions) to prevent redundant work. This adaptive classification operates without pre-defined category constraints.

\subsubsection{audit\_output: Quality Verification}

\paragraph{Masked Validation Protocol.} The \texttt{audit\_output\_tool} performs quality verification with pre-detection suppression. To prevent false positives from LVLM re-detection of already anonymized regions, the tool constructs a synthetic masked image $I_{\text{masked}}$ where all processed areas are replaced with black pixels. The LVLM classifier operates on $I_{\text{masked}}$ to identify residual PII instances $\mathcal{P}_{\text{residual}} = \text{LVLM}(I_{\text{masked}}, \theta_{\text{classify}})$. The function returns success status (true when $|\mathcal{P}_{\text{residual}}| = 0$), the list of detected residual instances, current iteration count, and a termination flag indicating whether maximum attempts ($t \geq 3$) were reached. This pre-detection suppression eliminates false positives on anonymized areas while the iteration limit prevents infinite retry loops.

\subsubsection{log\_output: Result Persistence}

The \texttt{log\_output\_tool} persists anonymization results and marks pipeline completion. Given the original image path and final anonymized output path, the tool saves result metadata and returns success status indicators.

\subsection{Tool Specifications: GenerativeAgent}

\subsubsection{anonymize\_and\_inpaint}

\paragraph{Algorithmic Design.} The anonymization function $\mathcal{T}_{\text{inpaint}}: \mathcal{I} \rightarrow I_{\text{out}}$ processes instance set $\mathcal{I}$ via diffusion-based inpainting with Canny edge guidance. For each instance $i_j \in \mathcal{I}$:

\begin{enumerate}
    \item \textbf{IoU-based deduplication}: Check overlap with processed instances (threshold $\tau = 0.3$), skip if redundant
    \item \textbf{Scout-and-zoom segmentation}: Apply Grounded-SAM-2 on cropped region (20\% margin) to obtain precise mask $m_j$
    \item \textbf{Coordinate mapping}: Transform $m_j$ to full-image coordinates as $M_j$
    \item \textbf{Diffusion inpainting}: Apply SDXL with Canny ControlNet conditioning at $768 \times 768$ resolution
\end{enumerate}

\paragraph{Configuration Parameters.} The diffusion pipeline operates with:
\begin{itemize}
    \item Denoising strength: $\alpha = 0.9$ (strong modification)
    \item Sampling steps: $T = 25$ (SDXL)
    \item ControlNet scale: $\lambda_{\text{canny}} = 0.3$ (moderate edge preservation)
    \item Guidance scale: $s = 9.0$
    \item Color matching: disabled ($\beta = 0.0$)
    \item IoU threshold: $\tau = 0.3$ (30\% overlap for deduplication)
\end{itemize}

\paragraph{Implementation.} 
The \\ \texttt{anonymize\_and\_inpaint\_tool} processes instances sequentially (not batched) through SDXL + Canny ControlNet anonymization. For each instance $i_j$, the tool first checks IoU overlap against all processed instances using configurable threshold \texttt{PII\_IOU\_THRESHOLD} (default $0.3$). Instances with high overlap are skipped with status \texttt{iou\_overlap\_with\_processed}, preventing redundant reprocessing when AuditorAgent detects the same region across iterations. For novel instances, scout-and-zoom segmentation crops region $R_j$ with 20\% margin, applies Grounded-SAM-2, and maps the resulting mask $m_j$ to full-image coordinates as $M_j$. SDXL then inpaints the masked region at $768 \times 768$ resolution. Failed segmentations are skipped with status \texttt{scout\_zoom\_failed\_no\_fallback}. The function returns counts of successfully processed, skipped (IoU overlap or segmentation failures), and failed (inpainting errors) instances, along with per-instance status details and overall completion indicator.

\subsection{Phase 1 Deterministic Tools}

\subsubsection{segment\_persons: YOLOv8m-seg Instance Segmentation}

\paragraph{Architecture.} We employ YOLOv8m-seg\footnote{\url{https://github.com/ultralytics/ultralytics}, Accessed October 25, 2025}, a medium-scale one-stage detector with instance segmentation capability. The model provides real-time performance while maintaining high accuracy for person detection.

\paragraph{Post-processing.} Masks use tight YOLO segmentation boundaries:
\begin{itemize}
    \item Morphological dilation: Disabled
    \item Effect: Precise mask boundaries from YOLO segmentation without expansion
    \item Rationale: SDXL inpainting handles edge blending without requiring dilated masks
\end{itemize}

\paragraph{Implementation.} The \texttt{segment\_persons\_tool} applies YOLOv8m-seg for person instance segmentation. The model uses a confidence threshold $\tau_c = 0.25$ balancing recall and precision for privacy protection. Detected person masks are used directly without morphological dilation, relying on YOLO's accurate segmentation boundaries. The function returns detection count, binary masks, bounding boxes, and confidence scores. YOLOv8m provides a practical balance between processing speed and accuracy for real-time person detection.

\subsubsection{anonymize\_and\_inpaint\_persons: Three-Stage Pipeline}

\paragraph{Pipeline Architecture.} Person anonymization employs a cascaded LVLM-LVLM-diffusion architecture to eliminate appearance correlation while preserving contextual plausibility:

\begin{enumerate}
    \item \textbf{Visual description} (Qwen2.5-VL-32B~\cite{qwen2vl}): Extract semantic attributes $\mathcal{A}_i = \{\text{pose}, \text{clothing}, \text{action}\}$ for each person $i$
    \item \textbf{Attribute diversification} (Qwen2.5-32B): Transform $\mathcal{A}_i \rightarrow \mathcal{A}'_i$ by sampling colors from palette $\mathcal{C}$ with uniform distribution
    \item \textbf{Conditioned inpainting} (SDXL~\cite{sdxl} + OpenPose): Generate anonymized person $I'_i$ using pose-preserved synthesis with ControlNet~\cite{controlnet} conditioning
\end{enumerate}

\paragraph{Color Palette.} $\mathcal{C}$ = $\{$gray, beige, navy, white, black, \\ brown, khaki, \ldots$\}$ contains 20 diverse colors with brightness modifiers $\{\text{light, dark, bright, faded, ...}\}$ for minimal collision probability (prompt: see below).

\paragraph{Diffusion Configuration.}
\begin{align}
\begin{split}
\text{Resolution} &: 768 \times 768 \\
\alpha \, (\text{strength}) &: 0.9 \\
T \, (\text{steps}) &: 25 \, (\text{SDXL persons}) \\
\lambda_{\text{OP}} \, (\text{OpenPose}) &: 0.8 \\
s \, (\text{guidance}) &: 9.0 \\
\beta \, (\text{color matching}) &: 0.0 \, (\text{disabled})
\end{split}
\end{align}

\paragraph{Implementation.} The \\ \texttt{anonymize\_and\_inpaint\_persons\_tool} executes three-stage person anonymization with attribute diversification. First, Qwen2.5-VL-32B extracts semantic attributes $\mathcal{A}_i$ (pose, clothing, action) for each person $i$. Second, Qwen2.5-32B-Instruct transforms $\mathcal{A}_i \rightarrow \mathcal{A}'_i$ by sampling colors from the diverse palette $\mathcal{C}$, decoupling description from diversification. Third, SDXL with OpenPose ControlNet generates anonymized person $I'_i$ using pose-preserved synthesis with the configuration above. Color matching is disabled ($\beta = 0.0$) to allow complete appearance transformation while OpenPose conditioning maintains pose realism and context consistency. The function returns anonymized count, output path, and processing time.

\subsubsection{detect\_traffic\_signs: YOLO-TS Specialized Detection}

\paragraph{Model Specification.} We employ YOLO-TS~\cite{yolo-ts} (YOLO for Traffic Signs), a YOLOv8-based detector fine-tuned on German Traffic Sign Detection Benchmark (GTSDB). The model specializes in detecting traffic signs across diverse scales and lighting conditions.

\paragraph{Configuration.}
\begin{itemize}
    \item Model: GTSDB\_best.pt (trained on German traffic signs)
    \item Resolution: $1024 \times 1024$ pixels
    \item Confidence threshold: $\tau_c = 0.2$
    \item Output: Bounding boxes converted to binary exclusion masks
\end{itemize}

\paragraph{Implementation.} The \texttt{detect\_traffic\_signs\_tool} employs YOLO-TS with a YOLOv8 backbone and GTSDB-specialized detection head, trained on the German Traffic Sign Detection Benchmark. Operating at $1024 \times 1024$ resolution with confidence threshold $\tau_c = 0.2$, the model prioritizes recall to capture all public signage and avoid false negatives. Detected bounding boxes are converted to binary exclusion masks for unified masking with other PII categories. The function returns detection count, binary masks, bounding boxes, and traffic sign type IDs.

\subsubsection{detect\_license\_plates: Custom YOLOv8s Training}

\paragraph{Model Development.} We trained a custom YOLOv8s detector on UC3M License Plates (UC3M-LP) dataset~\cite{ramajo2024dual} to achieve high-recall license plate detection. The model operates at $1280 \times 1280$ resolution at inference for improved small object detection.

\paragraph{Training Configuration.}
\begin{itemize}
    \item Architecture: YOLOv8s
    \item Dataset: UC3M-LP (1,580 train, 395 val images)
    \item Resolution: $1280 \times 1280$ pixels
    \item Hardware: Single L40S GPU, batch size 8, 100 epochs
    \item Performance: Precision=0.93, Recall=0.94, mAP$_{50}$=0.91, mAP$_{50\text{-}95}$=0.82
\end{itemize}

\paragraph{Inference Configuration.}
\begin{itemize}
    \item Resolution: $1280$px (training resolution)
    \item Confidence: $\tau_c = 0.05$ (ultra-low threshold maximizes recall for privacy)
    \item Rationale: Prioritize recall over precision to ensure no license plates remain visible
\end{itemize}

\subsubsection{blur\_license\_plates: Gaussian Blur Anonymization}

\paragraph{Blur Parameters.} License plates are anonymized via Gaussian blur to ensure text illegibility while maintaining image naturalness.

\begin{align}
I'(x,y) = \sum_{i,j \in \mathcal{N}} I(x+i, y+j) \cdot G_{\sigma}(i,j)
\end{align}

where $G_{\sigma}$ is a Gaussian kernel with $\sigma = 8$ pixels and kernel size $15 \times 15$.

\paragraph{Implementation.} The \texttt{blur\_license\_plates\_tool} applies Gaussian blur anonymization to detected license plates. Using a Gaussian kernel with standard deviation $\sigma = 8$ pixels and kernel size $15 \times 15$, the filter ensures complete text illegibility while maintaining visual naturalness compared to pixelation approaches. The smooth falloff of the Gaussian function produces less jarring transitions at mask boundaries. The function returns the count of blurred plates and output path.

\section{Qualitative Anonymization Examples}
\label{sec:qualitative}

\begin{figure}[ht]
    \centering
    \includegraphics[width=0.12\columnwidth]{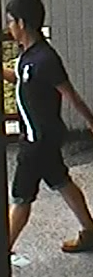}\hfill
    \includegraphics[width=0.12\columnwidth]{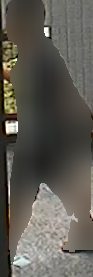}\hfill
    \includegraphics[width=0.12\columnwidth]{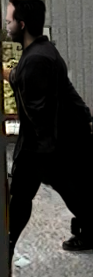}\hfill
    \includegraphics[width=0.12\columnwidth]{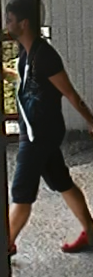}\hfill
    \includegraphics[width=0.12\columnwidth]{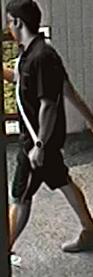}
    
    \vspace{0.1cm}
    
    \includegraphics[width=0.12\columnwidth]{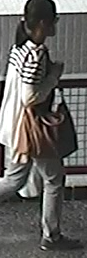}\hfill
    \includegraphics[width=0.12\columnwidth]{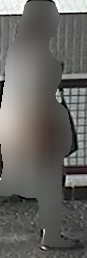}\hfill
    \includegraphics[width=0.12\columnwidth]{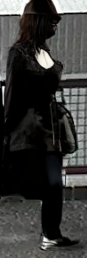}\hfill
    \includegraphics[width=0.12\columnwidth]{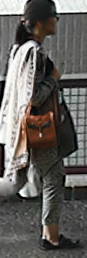}\hfill
    \includegraphics[width=0.12\columnwidth]{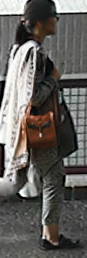}
    
    \vspace{0.1cm}
    
    \includegraphics[width=0.12\columnwidth]{figures/cuhk03_example3_original.png}\hfill
    \includegraphics[width=0.12\columnwidth]{figures/cuhk03_example3_blur.png}\hfill
    \includegraphics[width=0.12\columnwidth]{figures/cuhk03_example3_dp2.png}\hfill
    \includegraphics[width=0.12\columnwidth]{figures/cuhk03_example3_fadm.png}\hfill
    \includegraphics[width=0.12\columnwidth]{figures/cuhk03_example3_ours.png}
    
    \vspace{0.1cm}
    
    \includegraphics[width=0.12\columnwidth]{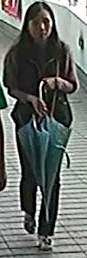}\hfill
    \includegraphics[width=0.12\columnwidth]{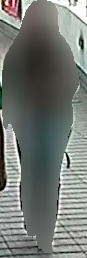}\hfill
    \includegraphics[width=0.12\columnwidth]{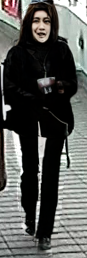}\hfill
    \includegraphics[width=0.12\columnwidth]{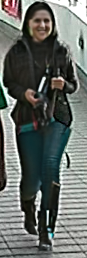}\hfill
    \includegraphics[width=0.12\columnwidth]{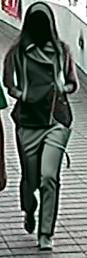}
    
    \vspace{0.1cm}
    
    \includegraphics[width=0.12\columnwidth]{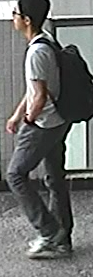}\hfill
    \includegraphics[width=0.12\columnwidth]{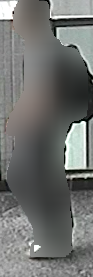}\hfill
    \includegraphics[width=0.12\columnwidth]{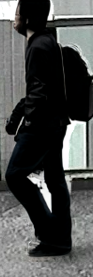}\hfill
    \includegraphics[width=0.12\columnwidth]{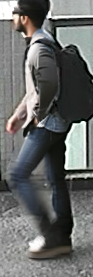}\hfill
    \includegraphics[width=0.12\columnwidth]{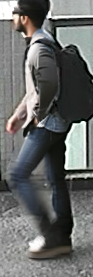}
    
    \vspace{0.05cm}
    
    \small
    \begin{tabular*}{\columnwidth}{@{\extracolsep{\fill}}ccccc@{}}
        Original & Blur & DP2 & FADM & Ours \\
    \end{tabular*}
    
    \caption{Qualitative comparison of person anonymization methods on CUHK03-NP test examples. Each row shows the same person processed by different methods (left to right): Original, Gaussian Blur, DeepPrivacy2 (DP2), FADM, and our approach. Our method preserves pose and scene structure while effectively anonymizing identities with photorealistic results. Blur destroys facial details and overall image quality. DP2 produces synthetic faces with visible artifacts. FADM maintains high similarity to originals (privacy risk). Our method achieves the optimal balance between privacy protection and visual quality.}
    \label{fig:cuhk03_examples}
\end{figure}

Figure~\ref{fig:cuhk03_examples} compares anonymization methods on CUHK03-NP test examples. Each row shows the same person processed by different techniques.

\section{Diffusion Model Conditioning}
\label{sec:models}

\subsection{Person Description Protocol}

\paragraph{LVLM Prompt Engineering.} 
We design the description prompt to extract semantic attributes while enforcing attribute diversification at the prompt level. 
The Qwen2.5-VL-32B model is prompted to describe a person’s build, pose, viewpoint, and action, while diversifying clothing attributes via randomized color and brightness selection. 
The operational prompt used for this purpose is defined as follows:

\begin{lstlisting}[language={},basicstyle=\ttfamily\tiny,breaklines=true]
DETECTOR_PERSON_DESCRIPTION_PROMPT = "You are describing a person for AI-based anonymization. Generate a SHORT description for Stable Diffusion inpainting.

CRITICAL: Invent DIVERSE clothing colors AND brightness - DO NOT copy from the image!
Colors (pick randomly): gray, beige, navy, white, black, brown, khaki, blue, red, green, yellow, pink, teal, burgundy, olive, charcoal, maroon, tan, cream, sage.
Brightness modifiers (pick randomly): light, dark, bright, faded, vibrant, muted, pale, deep, pastel, bold.
Example colors: 'light beige', 'dark navy', 'bright red', 'faded olive', 'deep burgundy', 'pale pink', 'vibrant teal'.

Include: body build (slim/medium/heavy), pose (standing/walking/sitting), view (front/side/back), action (talking/walking/standing).

SPECIAL CASE: If uniformed (police/fire/military/security), preserve uniform type and official colors.

Format: 'A [build] person [action], [view] view, wearing [brightness] [color] [top] and [brightness] [color] [bottom]'
Limit: 20-30 words maximum.

Examples:
1. Generic: {"description": "A medium-build person walking, side view, wearing dark olive jacket and pale charcoal pants"}
2. Generic: {"description": "A slim person standing, front view, wearing vibrant burgundy sweater and light khaki chinos"}
3. Generic: {"description": "A heavy-build person sitting, side view, wearing faded maroon hoodie and deep black jeans"}
4. Officer: {"description": "A police officer talking, facing forward, wearing navy police uniform with badge"}

Return ONLY valid JSON: {"description": "your text here"}"
\end{lstlisting}

\subsection{Person Inpainting Configuration}

\paragraph{SDXL Pipeline Parameters.} 
Person synthesis employs SDXL with OpenPose ControlNet. 
The positive and negative prompts (excluding the LVLM-generated description) are adapted from~\cite{FADM}, and the pipeline uses the following hyperparameters:

\begin{lstlisting}[language={},basicstyle=\ttfamily\tiny,breaklines=true]
# Prompt Construction
POSITIVE = "{LVLM_description}, RAW photo, 8k uhd, dslr, soft lighting, high 
quality, film grain, Fujifilm XT3, photorealistic, detailed"

NEGATIVE = "deformed iris, deformed pupils, semi-realistic, cgi, 3d, render, 
sketch, cartoon, drawing, anime, text, cropped, out of frame, worst quality, 
low quality, jpeg artifacts, ugly, duplicate, morbid, mutilated, extra fingers, 
mutated hands, poorly drawn hands, poorly drawn face, mutation, deformed, blurry, 
dehydrated, bad anatomy, bad proportions, extra limbs, cloned face, disfigured, 
gross proportions, malformed limbs, missing arms, missing legs, extra arms, 
extra legs, fused fingers, too many fingers, long neck"

# Hyperparameters
RESOLUTION       = 768        # Input/output resolution
STRENGTH         = 0.9        # Denoising strength α
STEPS            = 25         # Sampling steps T (SDXL persons)
GUIDANCE_SCALE   = 9.0        # Classifier-free guidance s
OPENPOSE_SCALE   = 0.8        # ControlNet conditioning strength λ_OP

# Color Matching (disabled for full appearance transformation)
LUMINANCE_MATCH  = 0.0
COLOR_STRONG     = 0.0
COLOR_SUBTLE     = 0.0
\end{lstlisting}

\paragraph{Justification.}
\begin{itemize}
    \item $\alpha = 0.9$: High denoising enables complete appearance transformation while preserving OpenPose structure.
    \item $T = 25$: SDXL architecture allows fewer steps than SD 1.5 while maintaining quality at 768px.
    \item $s = 9.0$: CFG scale for stable diffusion models.
    \item $\lambda_{\text{OpenPose}} = 0.8$: Balances pose preservation vs.\ natural variation.
\end{itemize}

\subsection{Object Inpainting Configuration}

\paragraph{SDXL Pipeline for Context-Dependent PII.} Object anonymization employs Canny edge conditioning for structure preservation with Stable Diffusion XL:

\begin{lstlisting}[language={},basicstyle=\ttfamily\tiny]
# Prompt Construction
POSITIVE = "{LVLM_description}"       # Use LVLM-provided semantic description
FALLBACK = "background scene"        # If LVLM description unavailable

NEGATIVE = ""                        # Empty by default (context-appropriate)

# Hyperparameters
RESOLUTION       = 768          # Input/output resolution
STRENGTH         = 0.9          # Denoising strength α
STEPS            = 25           # Sampling steps T (SDXL objects)
GUIDANCE_SCALE   = 9.0          # CFG scale s
CANNY_SCALE      = 0.3          # ControlNet conditioning λ_CN
CANNY_LOW        = 10           # Hysteresis low threshold
CANNY_HIGH       = 30           # Hysteresis high threshold

# Color Matching (disabled for complete transformation)
LAB_STRONG       = 0.0
LAB_SUBTLE       = 0.0
\end{lstlisting}

\paragraph{Parameter Selection.}
\begin{itemize}
    \item $T = 25$: SDXL architecture enables efficient generation with fewer steps than SD 1.5
    \item $\lambda_{\text{Canny}} = 0.3$: Moderate edge guidance allows flexible background reconstruction during object removal
    \item Canny thresholds $(10, 30)$: Captures salient edges while filtering noise
    \item Empty negative prompt: Maximizes generation flexibility for diverse object types
\end{itemize}

\section{License Plate Detection: Extended Analysis}
\label{sec:license_plates}

\subsection{Training Methodology}

\paragraph{Dataset Characteristics.} UC3M-LP provides 1,975 annotated images with diverse plate formats, viewing angles, and lighting conditions representative of real-world street imagery.

\paragraph{Model Architecture.} YOLOv8s (11.2M parameters) balances detection accuracy with inference speed, trained at 1280px resolution for 100 epochs on a single NVIDIA L40S GPU.

\paragraph{Training Configuration.} Key hyperparameters:
\begin{itemize}
    \item \textbf{Optimizer}: SGD with lr$_0$=0.01, momentum=0.937, weight decay=0.0005
    \item \textbf{Learning rate schedule}: Cosine annealing with warmup (3 epochs), final lr=0.0001
    \item \textbf{Loss weights}: box=7.5, cls=0.5, dfl=1.5 (emphasizes localization accuracy)
    \item \textbf{Batch size}: 8 with mixed precision (AMP) training
\end{itemize}

\paragraph{Augmentation Strategy.} We employ aggressive data augmentation to improve generalization:

\begin{itemize}
    \item \textbf{Mosaic} (prob=1.0): Combines 4 images into single training sample, enhances multi-scale learning
    \item \textbf{Copy-paste} (prob=0.3): Synthesizes challenging occlusion scenarios
    \item \textbf{MixUp} (prob=0.1): Linear interpolation between samples for regularization
    \item \textbf{Multi-scale training}: $[0.5, 1.5] \times$ base resolution for scale invariance
    \item \textbf{Color jitter}: HSV adjustments (h=0.015, s=0.7, v=0.4) for lighting invariance
    \item \textbf{Spatial augmentation}: Horizontal flip (prob=0.5), rotation, translation, and scaling
\end{itemize}

\begin{figure}[h]
\centering
\includegraphics[width=0.48\textwidth]{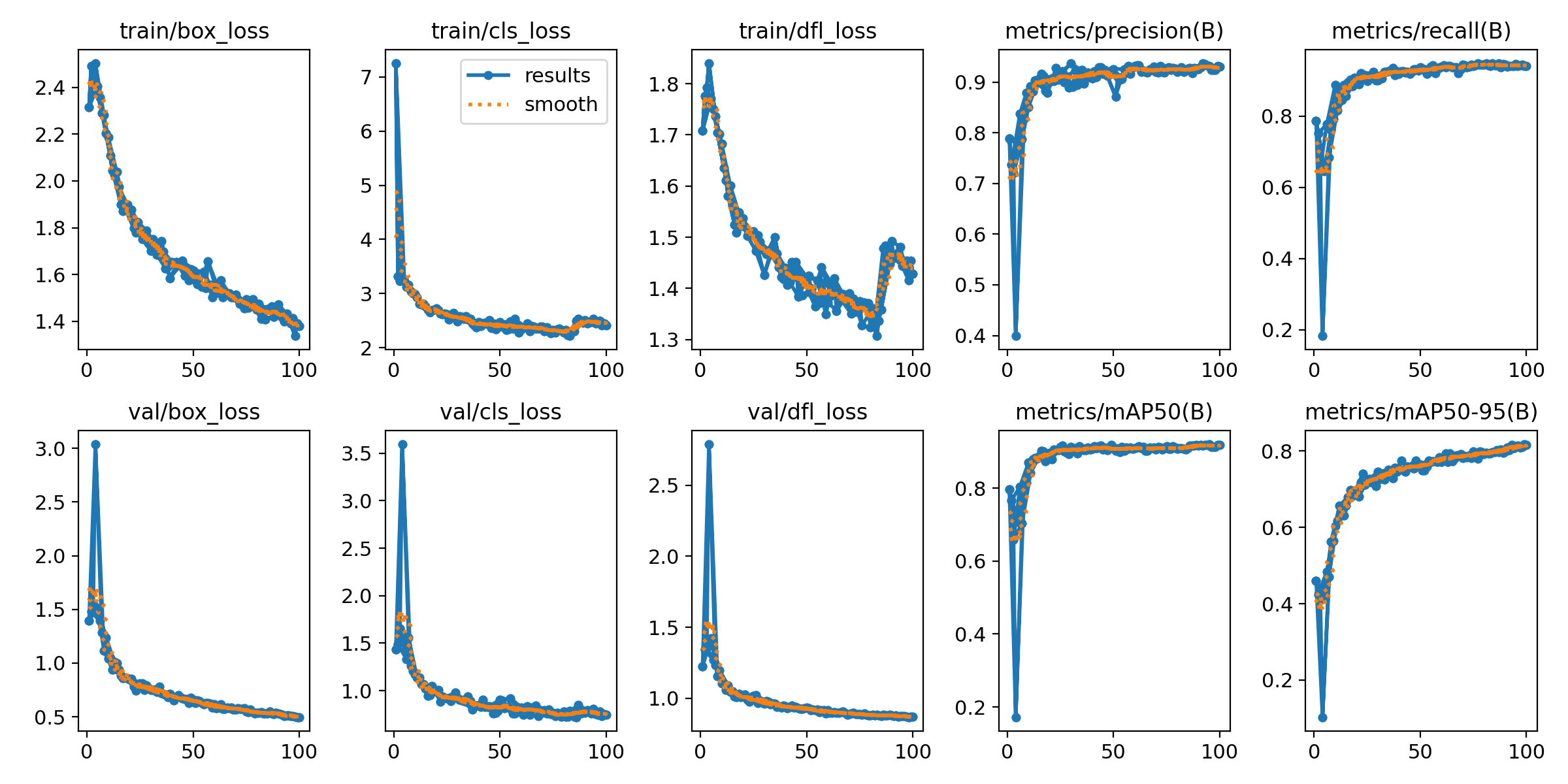}
\caption{\textbf{Training curves for YOLOv8s license plate detector.} Loss curves and metrics over 100 epochs. Stable convergence achieved by epoch 60.}
\label{fig:lp_training}
\end{figure}

\paragraph{Training Dynamics.} Figure~\ref{fig:lp_training} shows convergence behavior. The model achieves stable performance by epoch 60, with validation mAP$_{50\text{-}95}$ plateauing at 0.816. Training and validation losses demonstrate consistent convergence without overfitting, validating our augmentation strategy.

\begin{figure}[h]
\centering
\begin{minipage}{0.48\textwidth}
\centering
\includegraphics[width=\textwidth]{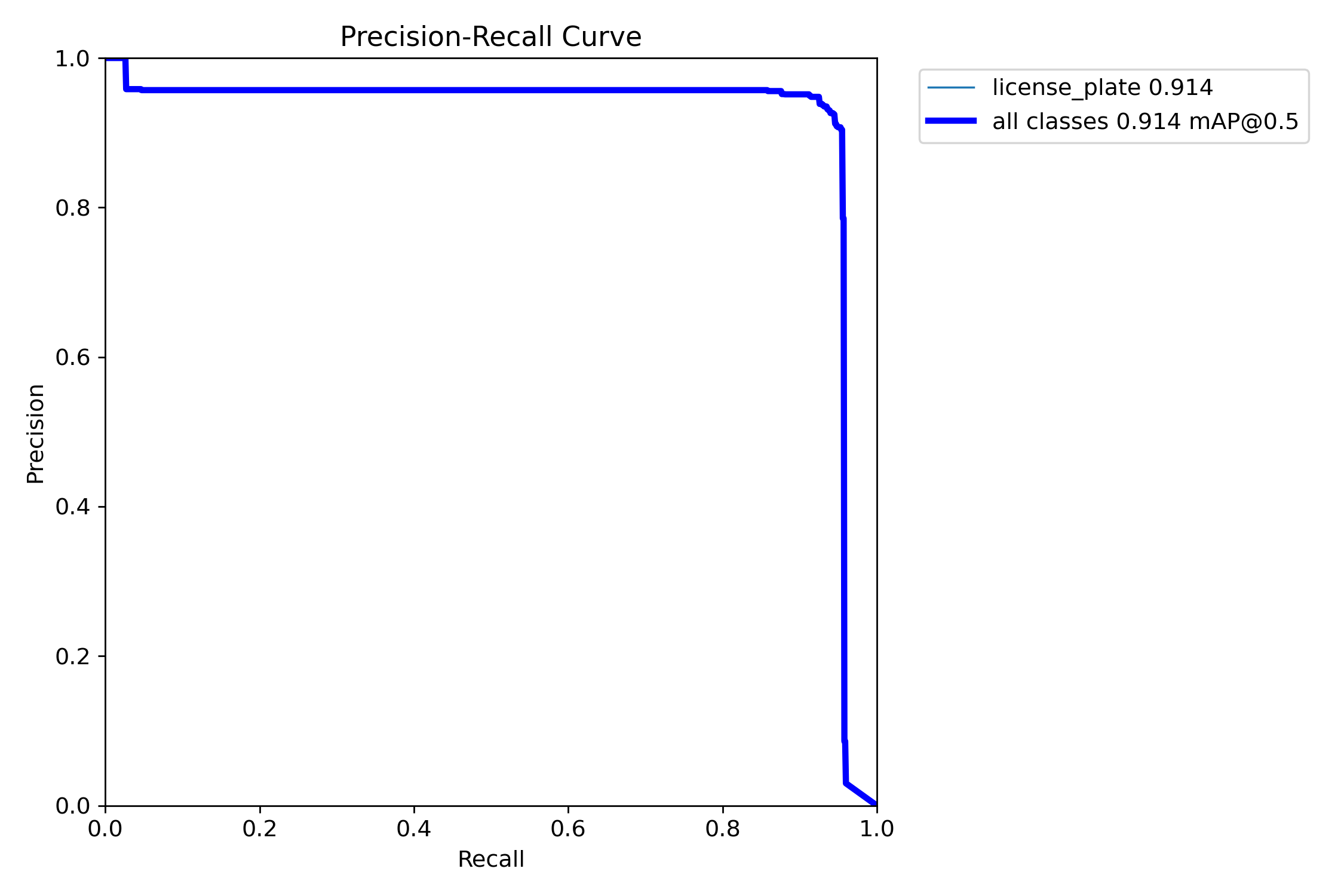}
\end{minipage}
\hfill
\begin{minipage}{0.48\textwidth}
\centering
\includegraphics[width=\textwidth]{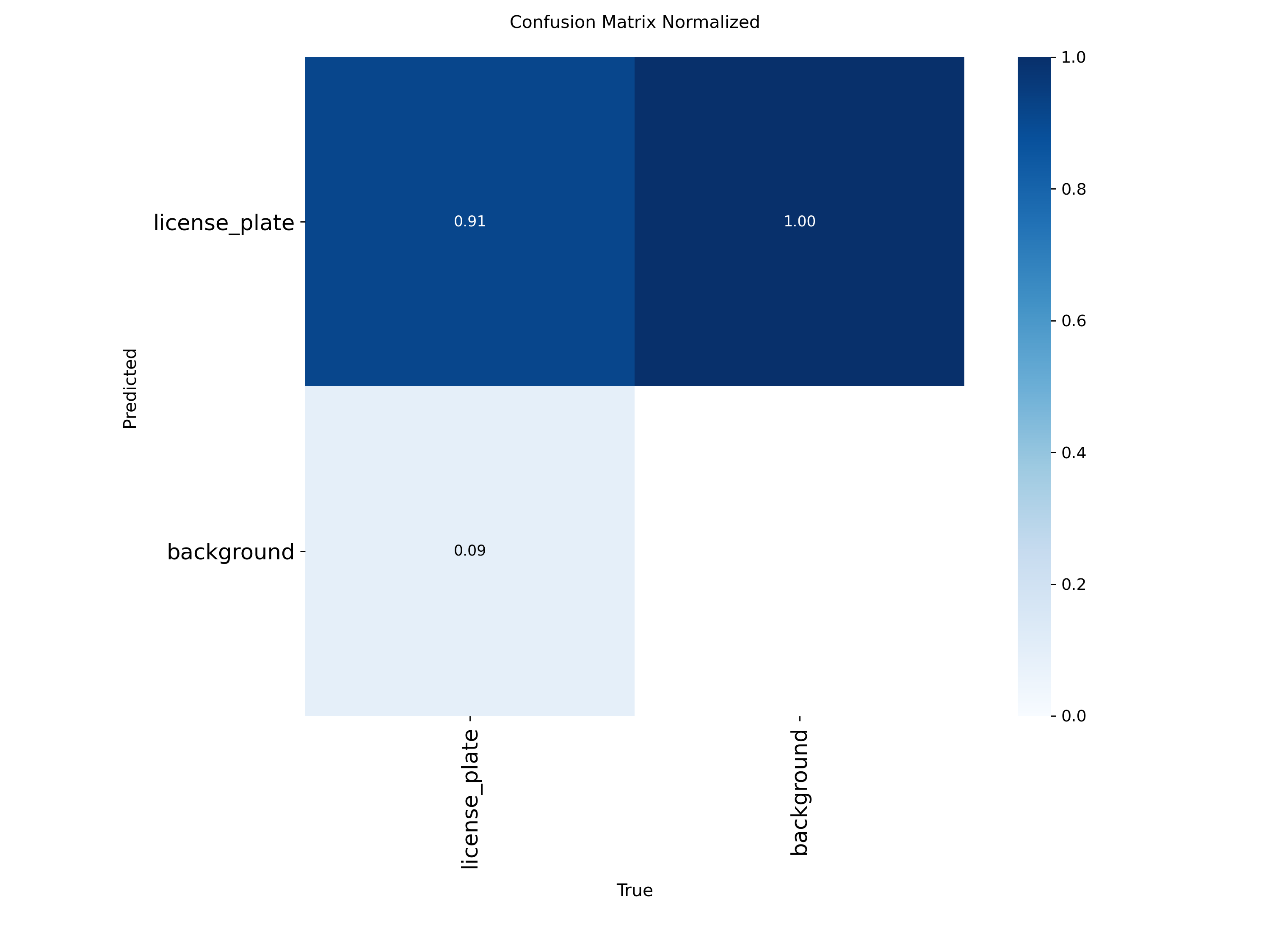}
\end{minipage}
\caption{\textbf{License plate detection performance.} Left: Precision-recall curve showing mAP$_{50}$=0.91. Right: Normalized confusion matrix with 0.94 true positive rate.}
\label{fig:lp_metrics}
\end{figure}

\paragraph{Performance Analysis.} Figure~\ref{fig:lp_metrics} presents precision-recall analysis. The model achieves:
\begin{align}
\begin{split}
\text{Precision} &= 0.93 \\
\text{Recall} &= 0.94 \\
\text{mAP}_{50} &= 0.91 \\
\text{mAP}_{50\text{-}95} &= 0.82
\end{split}
\end{align}

The high recall (0.94) is critical for privacy applications where false negatives are unacceptable. Precision-recall curve analysis shows robust performance across confidence thresholds.

\subsection{Training Configuration}

We train Mask R-CNN \cite{he2017mask} with ResNet-50-FPN \cite{lin2017feature} on Visual Redactions Dataset \cite{orekondy17iccv,orekondy2018redaction}: $3{,}873$ training images, $21{,}489$ annotations across $28$ privacy categories. COCO-pretrained weights initialize the backbone. Training uses SGD (momentum $0.9$, weight decay $0.0001$), base LR $0.002$ with warmup ($500$ iterations), decay at $18$K/$25$K iterations, batch size $8$ ($2\times$ L40S GPUs), $30$K iterations ($\sim$$5.7$ hours). Augmentation: random horizontal flip, multi-scale training ($640$--$800$px shortest edge).

\subsection{Results}

\begin{figure*}[t]
\centering
\begin{subfigure}[b]{0.48\textwidth}
    \includegraphics[width=\textwidth]{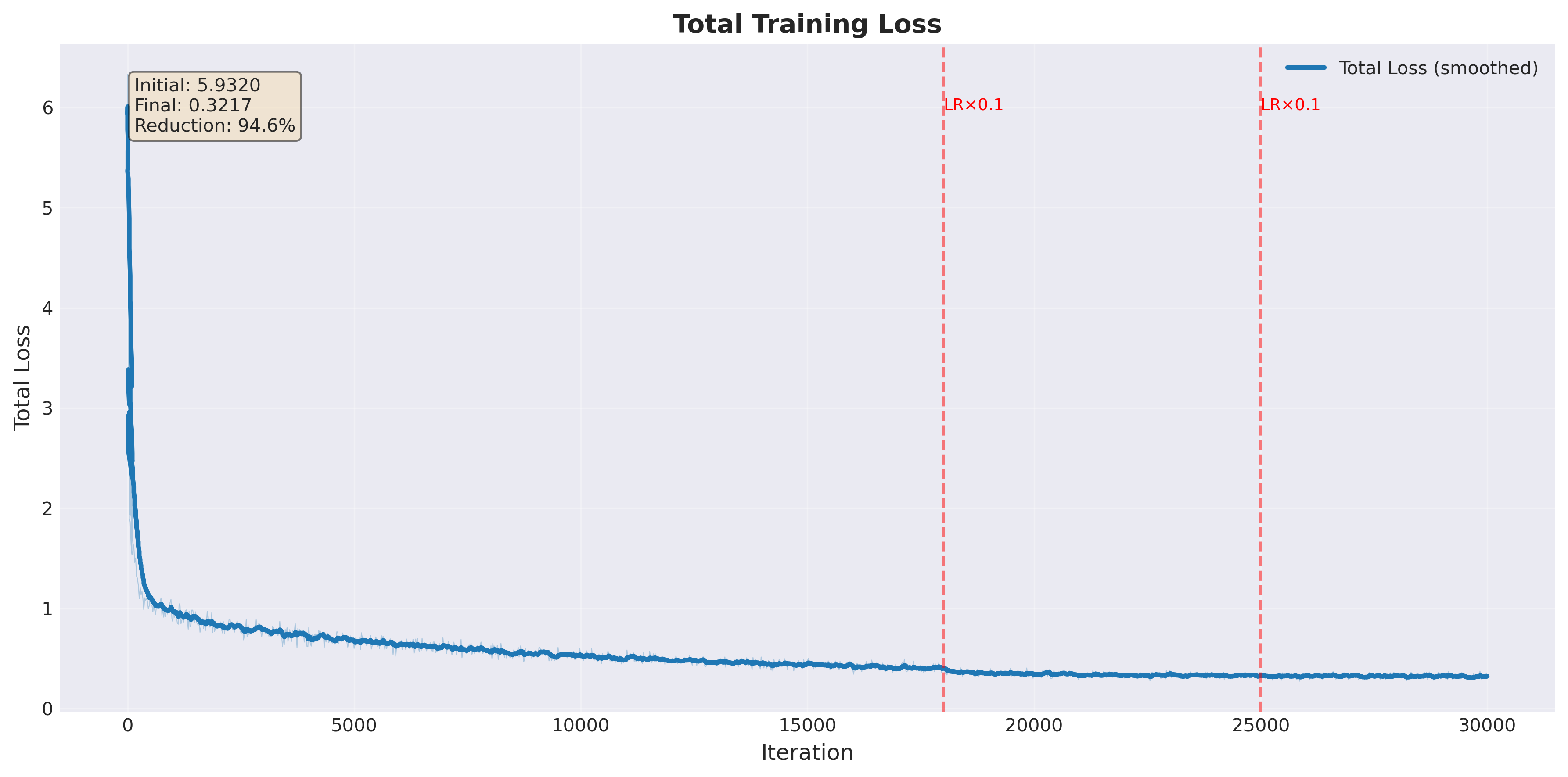}
    \caption{Total training loss}
\end{subfigure}
\hfill
\begin{subfigure}[b]{0.48\textwidth}
    \includegraphics[width=\textwidth]{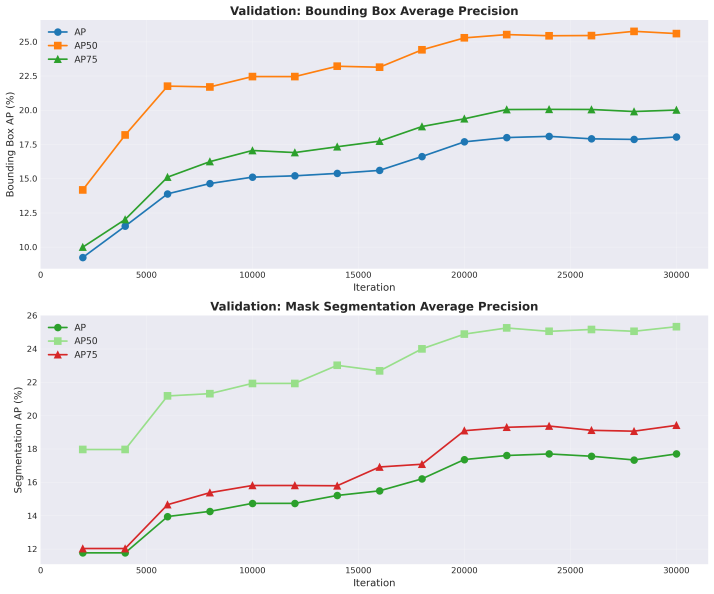}
    \caption{Validation AP progression}
\end{subfigure}

\begin{subfigure}[b]{0.48\textwidth}
    \includegraphics[width=\textwidth]{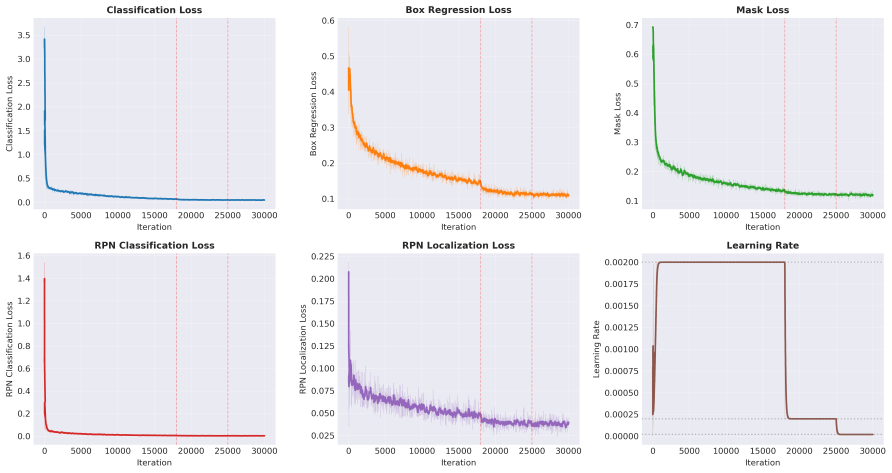}
    \caption{Component losses breakdown}
\end{subfigure}
\hfill
\begin{subfigure}[b]{0.48\textwidth}
    \includegraphics[width=\textwidth]{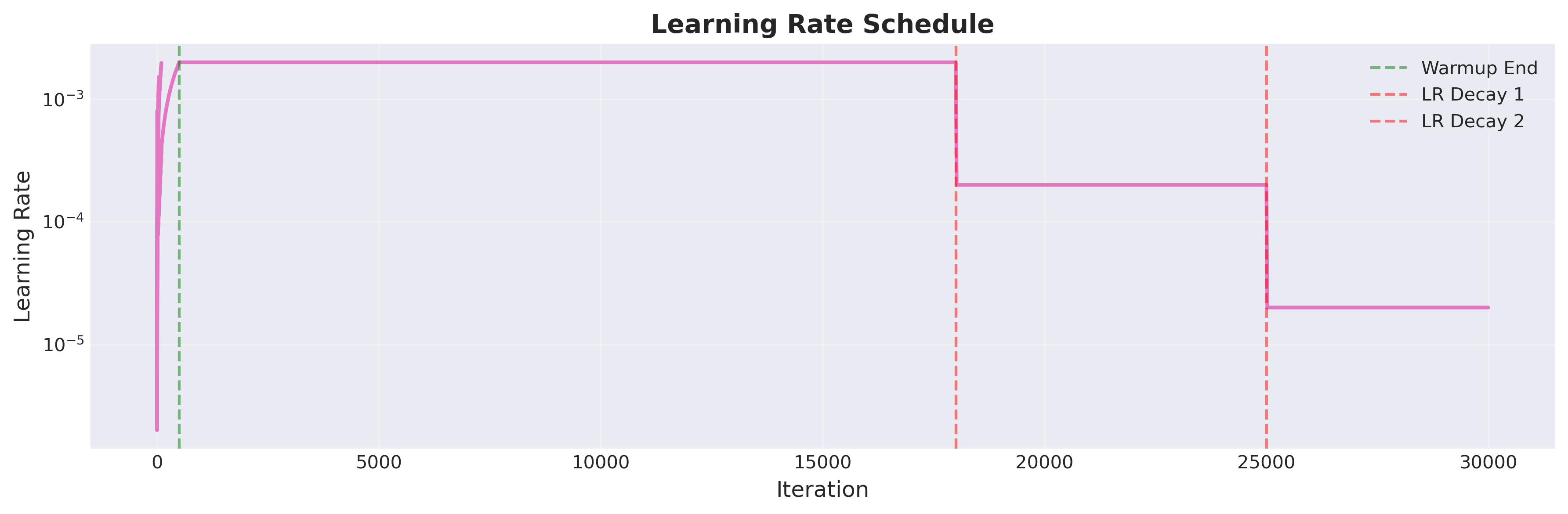}
    \caption{Learning rate schedule}
\end{subfigure}
\caption{\textbf{Training dynamics of Mask R-CNN on Visual Redactions Dataset.} (a) Total loss shows $94.6\%$ reduction over 30K iterations. (b) Validation AP: segmentation AP from $11.77\%$ to $17.70\%$, bbox AP from $9.24\%$ to $18.04\%$. (c) Component losses: RPN classification ($99.2\%$ reduction), classification ($98.5\%$), mask ($82.6\%$), box regression ($72.7\%$). (d) Learning rate schedule: linear warmup (500 iterations), decay ($\times0.1$) at 18K and 25K iterations.}
\label{fig:pii_seg_training}
\end{figure*}

\paragraph{Training Dynamics.} Total loss decreases from $5.93$ to $0.32$ over 30K iterations ($94.6\%$ reduction). Component-wise: RPN classification ($99.2\%$ reduction), classification loss ($98.5\%$), mask loss ($82.6\%$), box regression ($72.7\%$). Validation AP improves monotonically from $11.77\%$ (2K) to $17.70\%$ (30K) without overfitting. Figure~\ref{fig:pii_seg_training} shows training dynamics.

\paragraph{Category-Level Performance.} Overall mask AP: $17.70\%$ (AP$_{50}$: $25.33\%$, AP$_{75}$: $19.41\%$). Severe scale dependency: AP$_S$: $3.37\%$, AP$_L$: $19.60\%$. Performance clusters by attribute type (Table~\ref{tab:category_ap}): \textit{Visual classes} (face: $56.98\%$, person: $48.68\%$, passport: $70.55\%$) exploit COCO pretraining successfully. \textit{Textual categories} exhibit systematic failure ($10/10$ categories $<$$5\%$ AP: name $1.16\%$, address $0.64\%$, phone $0.00\%$), despite abundant data (address: $1{,}902$ instances, date\_time: $2{,}979$ instances). Small object scale (AP$_S$: $3.37\%$) causes detection failure on spatially compact text.

\paragraph{Test Set.} Unified binary masks (logical OR across 28 categories) achieve Dice: $75.83\%$, IoU: $68.71\%$, Precision: $80.71\%$, Recall: $78.02\%$ (2,890/2,989 images; 99 missing predictions). Performance dominated by visually salient categories; $\sim$$22\%$ missed pixels primarily from small text below detection thresholds.

\begin{table}[t]
\centering
\caption{Per-category mask AP on Visual Redactions validation set ($n=1{,}611$ images). Categories grouped by performance regime. Training instance counts reflect severe class imbalance. Abundant data does not guarantee success: date\_time ($2{,}979$ instances $\rightarrow$ $3.75\%$ AP), address ($1{,}902$ instances $\rightarrow$ $0.64\%$ AP). Category distribution shows visual attributes (face, person) dominate both instance count ($46.4\%$) and performance, while textual categories suffer systematic failure despite comprising $36\%$ of training instances.}
\label{tab:category_ap}
\small
\begin{tabular}{lrr}
\toprule
\textbf{Category} & \textbf{Instances} & \textbf{AP (\%)} \\
\midrule
\multicolumn{3}{l}{\textit{High Performance (AP $>$ 20\%)}} \\
Passport & 153 & 70.55 \\
Face & 4{,}039 & 56.98 \\
Person & 5{,}937 & 48.68 \\
Fingerprint & 79 & 45.02 \\
License Plate & 371 & 33.02 \\
Student ID & 37 & 32.35 \\
Ticket & 420 & 32.14 \\
Receipt & 113 & 30.35 \\
Credit Card & 102 & 18.06 \\
\midrule
\multicolumn{3}{l}{\textit{Medium Performance ($5\%$ $<$ AP $\leq$ $20\%$)}} \\
Disability & 73 & 17.76 \\
Drivers License & 30 & 16.16 \\
Nudity & 320 & 15.20 \\
Ethnic Clothing & 123 & 12.74 \\
Mail & 73 & 12.21 \\
Medicine & 116 & 10.50 \\
Handwriting & 1{,}077 & 10.29 \\
Birth Date & 60 & 7.48 \\
Signature & 252 & 7.40 \\
\midrule
\multicolumn{3}{l}{\textit{Low/Zero Performance (AP $\leq$ $5\%$)}} \\
Education & 48 & 5.18 \\
Date/Time & 2{,}979 & 3.75 \\
Username & 310 & 3.51 \\
Address Home & 90 & 2.11 \\
Landmark & 807 & 1.64 \\
Name & 1{,}785 & 1.16 \\
Email & 68 & 0.69 \\
Address & 1{,}902 & 0.64 \\
Phone & 105 & 0.00 \\
ID Card & 20 & 0.00 \\
\midrule
\textbf{Total} & \textbf{21{,}489} & \textbf{17.70} \\
\bottomrule
\end{tabular}
\end{table}

\section{Qualitative Comparison: Supervised vs. Zero-Shot}
\label{sec:qualitative_comparison}

\begin{figure*}[t]
\centering
\includegraphics[width=1.2\textwidth, angle=-90]{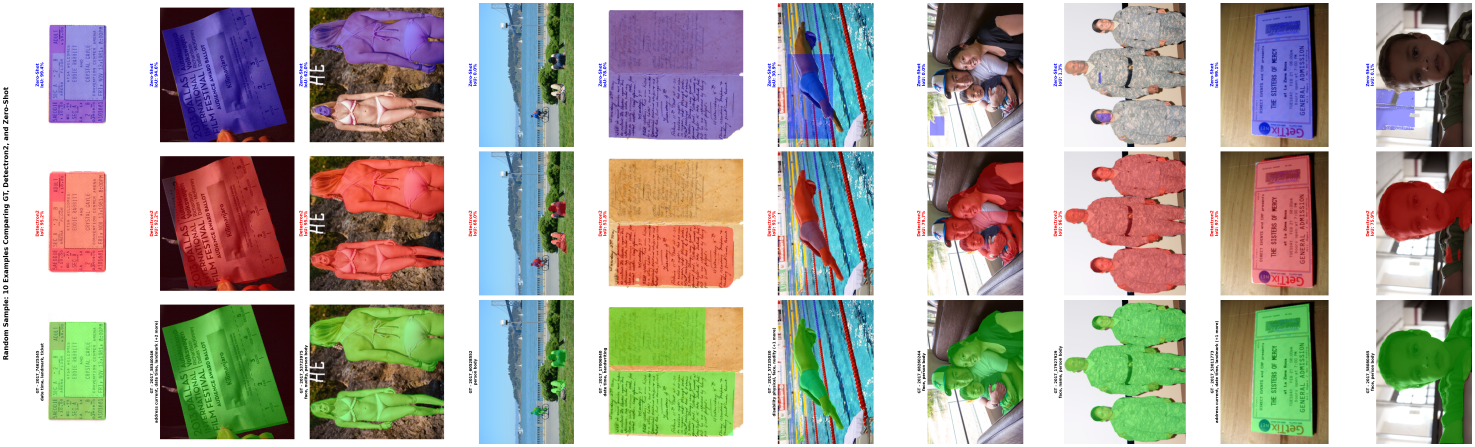}
\caption{\textbf{Random examples from Visual Redactions test set.} Ground truth (with category labels), Detectron2, and zero-shot predictions. Supervised achieves higher precision on common categories but fails on 99 images (3.3\%). Zero-shot provides full coverage with broader generalization but occasional false positives.}
\label{fig:random_examples}
\end{figure*}

Figure~\ref{fig:random_examples} shows randomly sampled test images comparing ground truth, Detectron2, and zero-shot predictions. Detectron2 achieves higher precision on faces, persons, and documents (categories with abundant training data and COCO initialization) but fails completely on 99/2,989 test images (3.3\%). Zero-shot provides broader category generalization but produces false positives on ambiguous regions.

\begin{figure*}[t]
\centering
\includegraphics[width=0.5\textwidth]{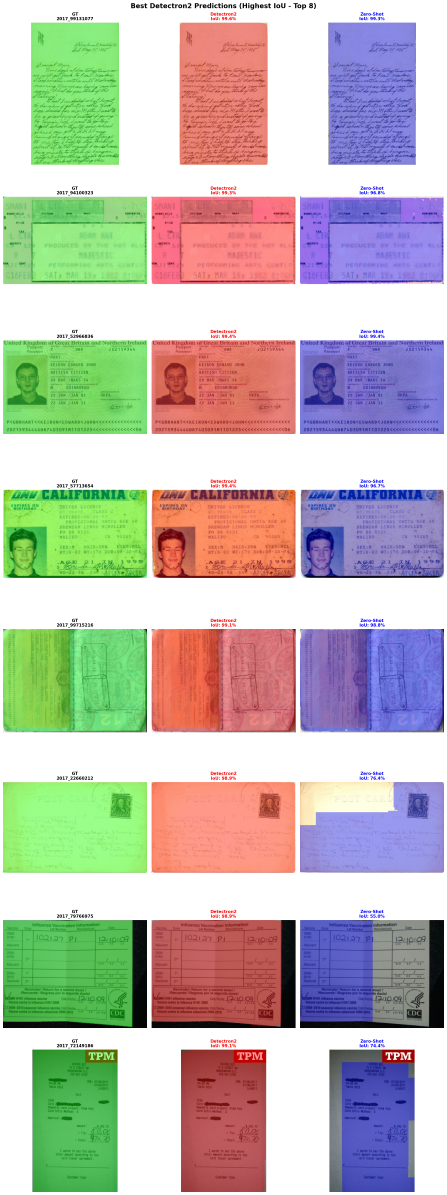}
\caption{\textbf{Detectron2 best cases.} Top 4 examples by Dice score. Supervised excels on faces, persons, documents with abundant training data and COCO initialization.}
\label{fig:best_detectron2}
\end{figure*}

\begin{figure*}[t]
\centering
\includegraphics[width=0.5\textwidth]{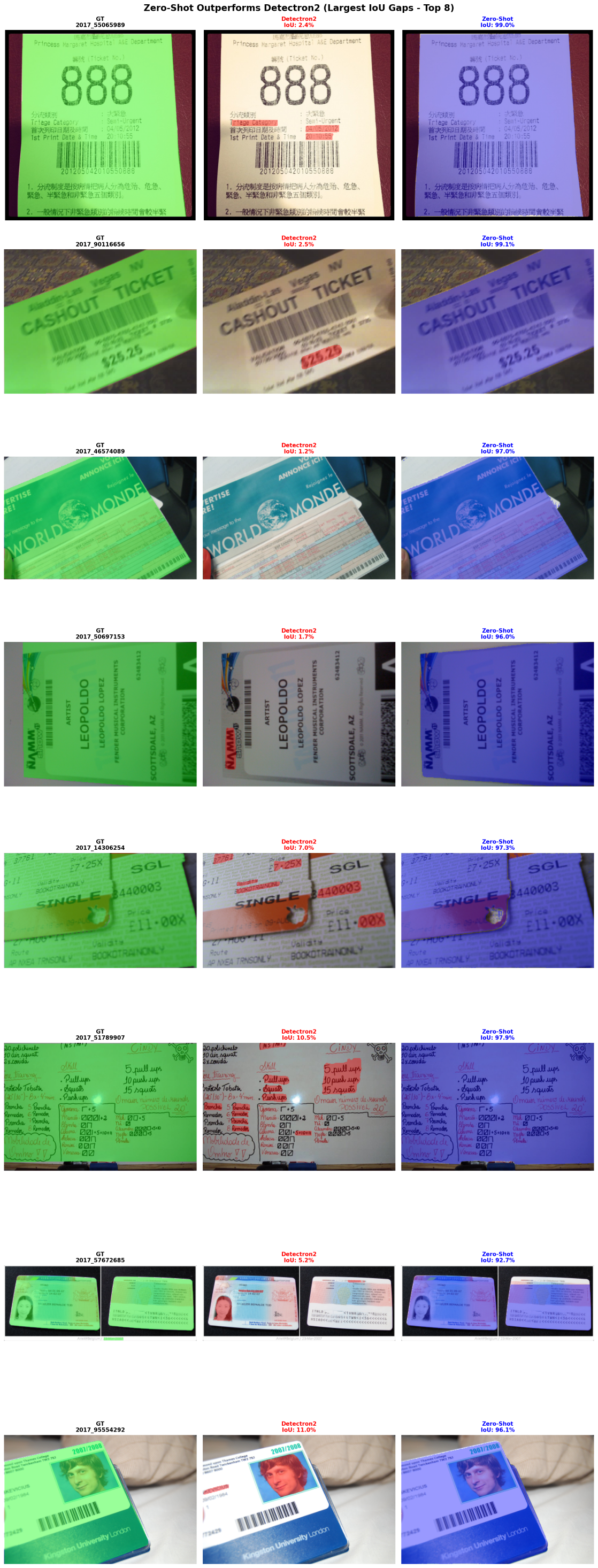}
\caption{\textbf{Zero-shot advantage cases.} Top 4 examples where zero-shot outperforms Detectron2. Agents with LVLM utilization succeed on rare categories, novel formats, and low-confidence regions.}
\label{fig:zeroshot_wins}
\end{figure*}

Figure~\ref{fig:best_detectron2} shows Detectron2 best cases: accurate mask boundaries on faces (AP: 56.98\%), persons (48.68\%), and passports (70.55\%). Figure~\ref{fig:zeroshot_wins} shows zero-shot advantages.

\section{KITTI Dataset Example}
\label{sec:kitti_example}

We processed KITTI~\cite{Geiger2012CVPR} image 0000000165 to evaluate cross-dataset generalization. KITTI provides street-level autonomous driving imagery at lower resolution than CityScapes ($2048\times1024$).

\begin{figure*}[t]
\centering
\includegraphics[width=0.45\textwidth]{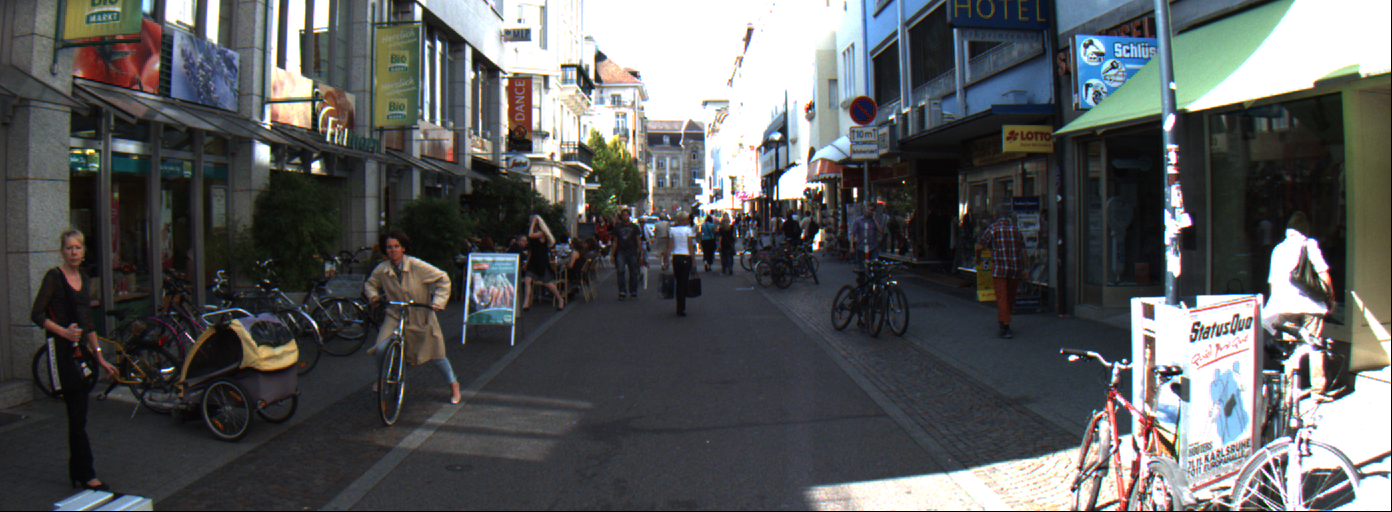}
\hfill
\includegraphics[width=0.45\textwidth]{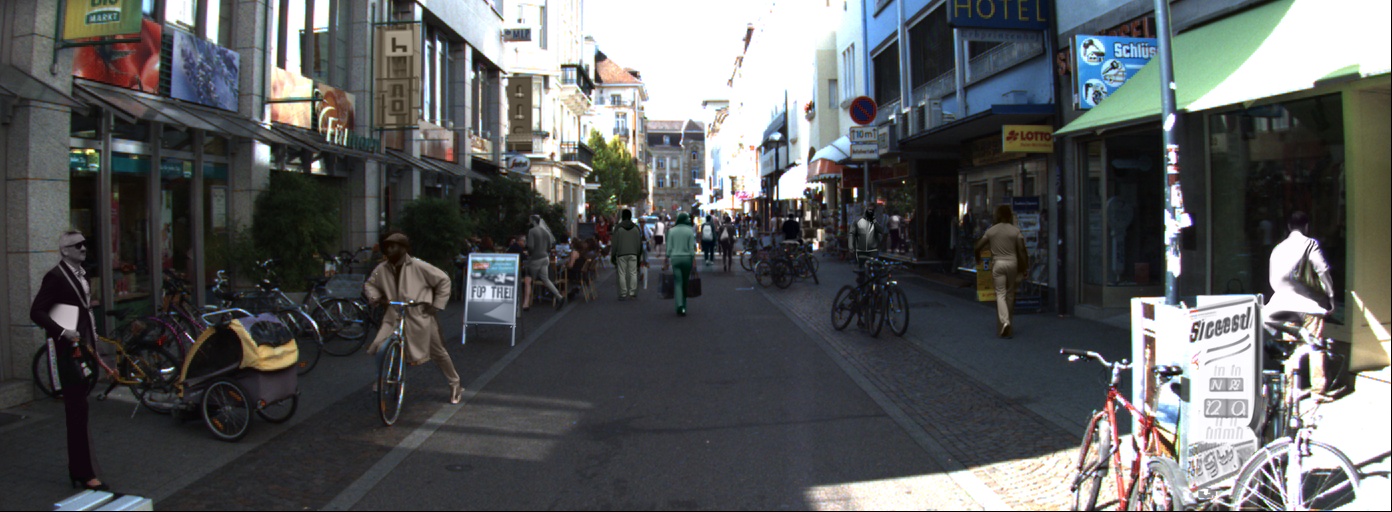}
\caption{\textbf{KITTI example (0000000165).} \textbf{Left:} Original street scene with 14 pedestrians and multiple traffic signs. \textbf{Right:} Anonymized output after PDCA workflow. Phase 1 batch-inpainted 14 persons using SDXL with OpenPose ControlNet. Phase 2 processed 3 rectangular signs with Canny ControlNet. Final PII coverage: 5.29\% (37,677 pixels). Total processing time: 391.8s with 3 PDCA iterations.}
\label{fig:kitti_example}
\end{figure*}

\paragraph{Detection and Processing.}
The pipeline detected 14 persons and 3 rectangular signs (Figure~\ref{fig:kitti_example}). Phase 1 batch-inpainted all persons using SDXL with OpenPose ControlNet. The LVLM generated appearance descriptions for close person bounding boxes: ``A medium-build person walking, front view, wearing muted navy blazer and light tan pants'', ``A medium-build person standing, side view, wearing bright yellow t-shirt and dark teal pants''. Phase 2 processed rectangular signs using Canny ControlNet.

\paragraph{PDCA Iteration and Audit.}
The workflow executed 3 PDCA iterations before reaching $n_{\max}=3$. The final audit detected 1 residual sign at bbox [948, 0, 1064, 100], likely a distant or partially occluded sign at the image boundary. This illustrates the privacy-utility trade-off: $n_{\max}$ bounds execution time (391.8s) while occasionally leaving residual PII.

\paragraph{Agent Dialogue Excerpt.}
Table~\ref{tab:kitti_dialogue} shows Phase 2 agent interactions. After Phase 1 completion, the AuditorAgent invokes \texttt{classify\_pii}, identifying 1 rectangular sign at bbox [464, 275, 526, 341] (instance ID: pii-0001). The OrchestratorAgent routes this to the GenerativeAgent, which invokes \texttt{anonymize\_and\_inpaint}. Tool execution includes scout-and-zoom segmentation (crop generation, Grounded-SAM-2 detection, mask mapping), protected area subtraction (15 masks from persons and traffic signs), and SDXL inpainting with Canny ControlNet. Final PII coverage: 5.29\% (37,677 pixels).

\begin{table}[t]
\centering
\caption{Agent dialogue excerpt from KITTI Phase 2 workflow (indirect PII detection and anonymization).}
\label{tab:kitti_dialogue}
\small
\begin{tabular}{p{2.5cm}p{5cm}}
\toprule
\textbf{Agent} & \textbf{Action/Message} \\
\midrule
UserProxyAgent & Phase 1 complete. AuditorAgent: classify PII in \texttt{0000000165.png} using \texttt{classify\_pii}. \\
\midrule
AuditorAgent & \textbf{Tool call:} \texttt{classify\_pii} \\
 & \texttt{image: artifacts/.../0000000165.png} \\
\midrule
Tool Result & \texttt{instances: [}\{\texttt{det\_prompt: "rectangular sign with text .", bbox: [464, 275, 526, 341], instance\_id: "pii-0001"}\}\texttt{]} \\
 & \texttt{meta: \{provider: ollama, model: qwen2.5vl:32b, tiles: 1\}} \\
\midrule
OrchestratorAgent & Phase 2: [classify\_pii $\checkmark$] Found indirect PII. GenerativeAgent, anonymize instances. \\
\midrule
GenerativeAgent & \textbf{Tool call:} \texttt{anonymize\_and\_inpaint} \\
 & \texttt{instances: [}\{\texttt{bbox: [464, 275, 526, 341], description: "rectangular sign with text", instance\_id: "pii-0001"}\}\texttt{]} \\
\midrule
Tool Execution & Scout-and-zoom: crop [464, 275, 526, 237] $\rightarrow$ detect $\rightarrow$ map mask \\
 & Protected area subtraction: 15 masks (persons, traffic signs) \\
 & Inpainting: SDXL + Canny ControlNet \\
 & PII mask update: 37,677 pixels (5.29\% coverage) \\
 ...
\end{tabular}
\end{table}

\paragraph{Statistics.}
The processing generated 14 person masks and 3 sign masks. Combined PII coverage: 5.29\% (37,677 pixels), primarily from pedestrians (4.88\% persons, 0.41\% signs). Processing time: 391.8s total — Phase 1: 160.5s for Grounded-SAM-2 detection; Phase 2: 231.3s for 3 PDCA iterations with Qwen2.5 VL 32B auditing and Qwen2.5 32B agent reasoning. Each detection included interpretable tool calls and bbox annotations.

\paragraph{Cross-Dataset Generalization.}
The KITTI example tests cross-dataset generalization: lower resolution than CityScapes ($2048\times1024$), different camera perspective (dashboard vs. urban mapping vehicle), and varying scene complexity. Zero-shot detection and pose-guided inpainting adapted to KITTI without dataset-specific tuning.

\end{document}